\documentclass[letterpaper]{article} 
\usepackage{aaai24}  
\usepackage{times}  
\usepackage{helvet}  
\usepackage{courier}  
\usepackage[hyphens]{url}  
\usepackage{graphicx} 
\usepackage{natbib}  
\usepackage{caption} 
\usepackage{algorithm}
\usepackage{algorithmic}

\usepackage{multirow}
\usepackage{subfig}
\usepackage{float}
\usepackage{multicol}

\usepackage{amsmath}
\usepackage{amsthm}
\usepackage{booktabs}
\usepackage{algorithm}
\usepackage{algorithmic}
\usepackage[switch]{lineno}
\usepackage{multirow}
\usepackage{graphicx}
\usepackage{subfig}
\usepackage{xcolor}
\usepackage{amsfonts}
\usepackage[scientific-notation=true]{siunitx}

\usepackage{newfloat}
\usepackage{listings}
\DeclareCaptionStyle{ruled}{labelfont=normalfont,labelsep=colon,strut=off} 
\floatstyle{ruled}
\newfloat{listing}{tb}{lst}{}
\floatname{listing}{Listing}
\title{Reprogramming Foundational Large Language Models(LLMs) for Enterprise Adoption for Spatio-Temporal Forecasting Applications: Unveiling a New Era in Copilot-Guided Cross-Modal Time Series Representation Learning}
\author{
    Sakhinana Sagar Srinivas\textsuperscript{\rm 1}\thanks{Designed, programmed the software, and drafted manuscript.},
    \{Chidaksh Ravuru\textsuperscript{\rm 2}, Geethan Sannidhi\textsuperscript{\rm 3}\}\thanks{Conducted experiments and analyzed visual results}, \\
    Venkataramana Runkana\textsuperscript{\rm 1}\\
}
\affiliations{
    \textsuperscript{\rm 1}TCS Research,
    \textsuperscript{\rm 2}IIT Dharwad,
    \textsuperscript{\rm 3}IIIT Pune \\

    \texttt{sagar.sakhinana@tcs.com}, \texttt{geethan.iiitp.ac.in}, \texttt{200010046@iitdh.ac.in} \\
    \texttt{venkat.runkana@tcs.com}\\
}

\usepackage{bibentry}

\begin{document}

\maketitle

\begin{abstract}
\vspace{-2mm}
Spatio-temporal forecasting plays a crucial role in various sectors such as transportation systems, logistics, and supply chain management. However, existing methods are limited by their ability to handle large, complex datasets. To overcome this limitation, we introduce a hybrid approach that combines the strengths of open-source large and small-scale language models (LLMs and LMs) with traditional forecasting methods. We augment traditional methods with dynamic prompting and a grouped-query, multi-head attention mechanism to more effectively capture both intra-series and inter-series dependencies in evolving nonlinear time series data. In addition, we facilitate on-premises customization by fine-tuning smaller open-source LMs for time series trend analysis utilizing descriptions generated by open-source large LMs on consumer-grade hardware using Low-Rank Adaptation with Activation Memory Reduction (LoRA-AMR) technique to reduce computational overhead and activation storage memory demands while preserving inference latency. We combine language model processing for time series trend analysis with traditional time series representation learning method for cross-modal integration, achieving robust and accurate forecasts. The framework's effectiveness is demonstrated through extensive experiments on various real-world datasets, outperforming existing methods by significant margins in terms of forecast accuracy.
\vspace{-5mm}
\end{abstract}

\vspace{-3mm}
\section{Introduction}
\begin{figure}[ht!]
\centering
\resizebox{1.1\linewidth}{!}{ 
\hspace*{-15mm}\includegraphics[keepaspectratio,height=4.5cm,trim=0.0cm 3.9cm 0cm 3.45cm,clip]{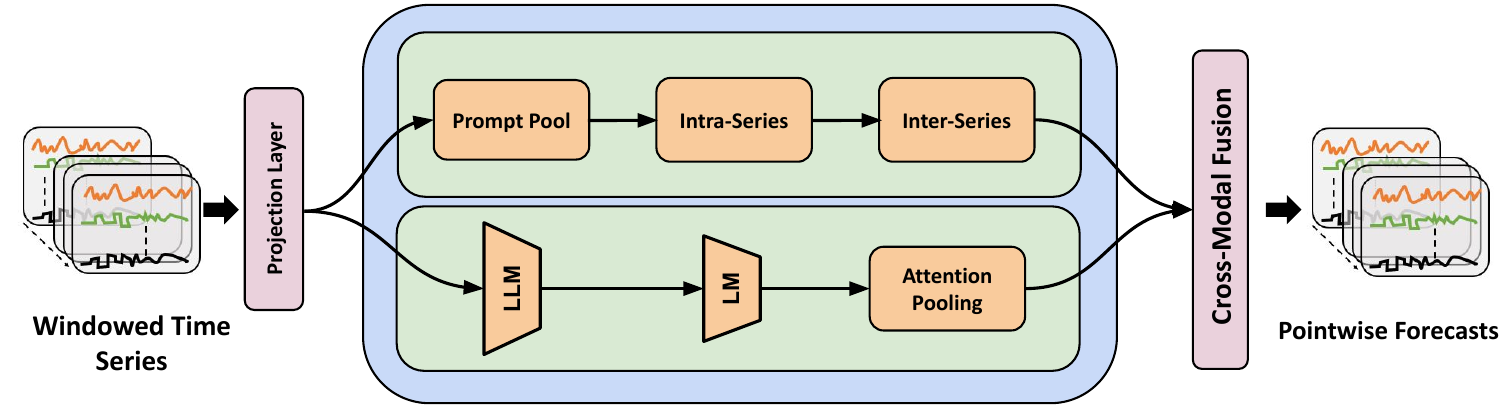} 
}
\vspace{-6mm}
\caption{Our framework incorporates the joint optimization of three methods: (a) a sequential stack of dynamic prompting mechanism to transfer relevant historical knowledge for adapting to new trends, coupled with learning intra- and inter-series dependencies to obtain contextualized time-series embeddings; (b) the utilization of large-scale model descriptions on time series trends to fine-tune a smaller language model, which then generates text-level embeddings encapsulating these trends; and (c) an output layer modeled with the multi-head attention (MHA) mechanism for integrating cross-domain embeddings and facilitating time series forecasting. This joint optimization framework provides a comprehensive and robust approach to modeling and forecasting spatio-temporal MTS data, enhancing adaptability, accuracy, and efficiency, and is designed for better generalization and scalability in real-world forecasting tasks.}
\label{fig:OverallArch}
\vspace{-2mm}
\end{figure}

\vspace{-1mm}
Multivariate time series forecasting (MTSF) is a long-standing task that finds wide application in various domains, enabling strategic decision-making through the prediction of multiple related variables that change over time. MTSF boasts numerous applications across various sectors with significant financial or operational impacts, such as transportation systems for route planning and navigation, logistics, and supply chain management for demand forecasting. However, MTSF presents several challenges, including complex relationships between time series variables, non-linearity, heterogeneity, sparsity, and non-stationarity. Recently, Spatio-Temporal Graph Neural Networks (STGNNs) have been introduced to improve multi-horizon forecast accuracy on multivariate time series (MTS) data. STGNNs are designed to accurately model both the long-range temporal dependencies within each individual variable time series and complex inter-dependencies among variables within the MTS data. STGNNs utilize both (i) explicit relationships, which are derived from predefined graphs created by human experts based on their prior domain knowledge, and (ii) implicit relationships, which are discovered through data-driven neural relational inference methods. Explicit relationships are static, often incomplete or inaccurate, while implicit relationships can exhibit significant non-linearity and evolve over time, revealing hidden connections among variables that may not be readily apparent to human experts. `Human-in-the-loop' STGNNs \cite{yu2017spatio, li2017diffusion, guo2020optimized} incorporate explicit knowledge but overlook the latent inter-relationships among the time series variables underlying the MTS data. A newer class of `human-out-of-the-loop' STGNNs \cite{deng2021graph, wu2020connecting, kipf2018neural} jointly infer the dependency graph structure, capturing the latent time-conditioned underlying relationships that drive the variable co-movements, while simultaneously learning the spatio-temporal dynamics from the MTS data to forecast future values. Despite their effectiveness, these approaches fail to fully leverage the accurate and reliable explicit (predefined) graph structures provided by domain experts, particularly when the data is noisy, leading to suboptimal forecasting. Transformers\cite{vaswani2017attention}, which excel in sequence modeling, are often preferred over STGNNs for interpreting spatio-temporal dynamics due to their ability to capture long-range dependencies through dynamic, context-aware self-attention mechanisms, offering greater scalability and flexibility. Furthermore, while existing STGNNs primarily focus on providing pointwise forecasts, they lack the ability to generate reliable uncertainty estimates for these forecasts. Uncertainty estimates are crucial for accurate risk assessment and informed decision-making. Moreover, current methods typically employ past data within a predetermined window length to learn historical patterns and predict future outcomes, with the duration of these patterns varying across different historical periods. Existing graph-based forecasting methods often rely on a fixed window length, adopting a one-size-fits-all approach. This approach may not be optimal as there is no universally ideal window length for all MTS data. In real-world applications, adjusting a predefined window length can be challenging, often unattainable, and computationally expensive. Based on our review of prior research, no existing methods fully capture the complexity of diverse patterns in MTS data, each characterized by varying lengths and distinct features. This makes achieving this goal an ambitious undertaking. In recent years, large language models (LLMs) such as GPT-4 \cite{gpt4} have revolutionized natural language processing(NLP), achieving remarkable performance by generating human-like responses and demonstrating enhanced logical reasoning and multitasking capabilities. These proprietary LLMs have acquired a vast and diverse range of linguistic constructs and knowledge through extensive pretraining on massive datasets. However, their internal workings remain largely opaque, earning them the moniker of `black-box' models. This lack of interpretability poses challenges for downstream applications, as these models typically do not provide direct access to logits or token embeddings. Furthermore, their `jack-of-all-trades' approach to handling a multitude of tasks often leads to suboptimal performance on specialized tasks. In contrast, open-source LLMs like Llama 2\cite{touvron2023llama} from Meta AI offer fine-tuning capabilities but necessitate substantial computational resources for adaptation to new tasks and domains through fine-tuning, primarily due to their large model sizes and the requirement for specialized hardware. Conversely, open-source small-scale language models (LMs) such as BERT\cite{devlin2018bert} are cost-effective for fine-tuning to specialized tasks using task-specific data and provide interpretability through access to logits or token embeddings. However, they may fall short in reasoning and generalization abilities, often generating less coherent and contextually relevant responses compared to larger LMs. Despite the transformative impact of LLMs in various domains, their application in time series analysis remains limited, primarily attributed to the scarcity of extensive datasets necessary for training LLMs for time series tasks. The largest publicly available datasets for time series analysis\cite{godahewa2021monash}, are significantly smaller than those employed in NLP tasks. While advancements have been made in utilizing LLMs, such as GPT-4, across various scientific disciplines, the synergistic integration of general-purpose LLMs with traditional forecasting methods for MTSF task remains an underexplored area in the development and advancement of intelligent forecasting techniques. This integration holds promise for achieving more accurate and robust future estimates. It is crucial to acknowledge that while this approach is innovative, LLMs such as GPT-4 are not inherently designed for time series analysis. Nonetheless, adapting them for this purpose, while unconventional, is entirely feasible. Originally conceived for NLP tasks, LLMs can be tailored for time series data, providing a unique method to generate comprehensive textual summaries that capture the main trends, patterns, and anomalies. These technical descriptions, encompassing trend analysis and data summarization, offer key insights and a multi-modal perspective that could complement traditional forecasting techniques. However, there is a significant limitation in sharing sensitive data with external LLM services. This includes the risks associated with sending regulated data to external LLM APIs to generate textual summaries on time series analysis. While LLMs offer a wide range of potential benefits for enterprises, their adoption is hindered by several limitations, including data privacy and sovereignty concerns, costs and customization requirements, and security vulnerabilities. To address these limitations, a novel approach termed `On-Premise Secure LLMs' is proposed. This solution would enable enterprises to fine-tune open-source large-scale LMs on their own proprietary data within their own infrastructure, enhancing data privacy and sovereignty, reducing costs, increasing customization options, and bolstering security. Overall, it offer a promising solution to the limitations of existing proprietary LLMs, potentially democratizing access to LLM capabilities and accelerating their adoption across a wide range of MTSF tasks, aligning with the growing demand for private, tailored AI solutions. Despite the advantages, on-premise LLMs face challenges such as high computational resource demands, scalability issues requiring extensive infrastructure upgrades, and the need for specialized technical expertise for deployment and maintenance. In this study, we introduce a novel framework built upon cross-modal time series representation learning, referred to as \texttt{LLM-TS Net} for brevity. The Figure \ref{fig:OverallArch} illustrates the proposed framework. The objective is to utilize the complementary strengths of open-source large and small-scale language models, and traditional forecasting methods to establish a more robust and accurate predictive framework. This approach models the time-varying uncertainty of framework predictions on future estimates, aiming to improve the accuracy of risk assessments and assist decision-makers by estimating predictive uncertainty. It presents a dynamic and flexible prompt mechanism designed to encode knowledge about various temporal dependencies and trends, and allows the method to adapt to the evolving nature of MTS data through transfer and reuse of learned knowledge and improve forecasting performance. This addresses the limitation of fixed, predefined window lengths, which often fail to account for the non-stationary nature of real-world MTS data. The framework adopts a time-then-space (TTS) approach, capturing the non-linear, temporal dynamics within times series variables before modeling the dependencies among different variables, offering a holistic understanding of MTS data. This is achieved through the integration of Grouped-query Multi-head Attention (GQ-MHA) for both intra- and inter-series analysis in spatio-temporal MTS data. The framework introduces a secure on-premise LLM based on the open-source, pretrained `llama2 7B 4k' model, customized for time series analysis and designed to run on consumer-grade hardware (low-cost GPUs). This approach not only reduces costs but also eliminates the need to transmit sensitive data to external servers, creating a secure and efficient environment that protects data privacy and offers customized intelligence without high costs or data sovereignty risks. We use custom prompts with task-specific instructions to query LLMs in a zero-shot setting, such as the `llama2 70B 4k' model. This approach enables us to generate textual descriptions that cover various aspects, such as identifying main trends, patterns, and requires the large-scale model to generalize and apply the implicit knowledge acquired during pretraining on vast text corpora for generating the desired output in analyzing MTS data. Next, we fine-tune small-scale LMs such as the `llama2 7B 4k' using the generated textual descriptions for task-specific adaptation to encapsulate the rich domain-specific knowledge within these descriptions. We utilize the open-source `llama2 70B 4k' model to analyze time series trends, owing to its advanced reasoning and inference capabilities, which are superior in handling complex tasks and yield more accurate and relevant textual descriptions compared to the `llama2 7B 4k' model. We then fine-tune the `llama2 7B 4k' model on the supervised task of minimizing cross-entropy loss, using pairs of MTS data-generated textual descriptions. This approach incorporates the rich, domain-specific insights extracted by the large-scale model from the MTS data. As a result, these smaller models become better equipped to handle similar tasks and are more aligned with the specific requirements of MTS data analysis.

\vspace{-4mm} 
 
\vspace{-2mm}
We introduce Low-Rank Adaptation with Activation Memory Reduction (LoRA-AMR), a powerful new tool for fine-tuning LLMs such as the `llama2 7B 4k' on generated descriptions for time series analysis. LoRA-AMR is efficient, fast, accurate, and enables new applications in various domains. This memory-efficient adaptation technique reduces trainable parameters and activation memory to update the low-rank weights, resulting in no additional computational overhead during fine-tuning and no increased latency during inference. It achieves comparable performance to full-parameter fine-tuning across various datasets on forecasting tasks. Furthermore, LoRA-AMR significantly cuts memory costs, allowing for the use of less powerful hardware. In contrast, the standard LoRA approach\cite{lora} is disadvantaged by its high activation memory consumption, which neither reduces nor may even increase the memory footprint compared to full-parameter fine-tuning, potentially creating a new memory bottleneck. Overall, this work presents the following contributions:

\vspace{-1mm}
\begin{itemize}
\item We present a dynamic prompting mechanism to enhance the adaptability and accuracy of time series representation learning methods. This mechanism recognizes and applies learned patterns from historical data, enabling it to adapt to changing data distributions. It functions similarly to knowledge transfer, applying previously learned patterns to new, similar input data. Additionally, the traditional method learns both intra- and inter-series dependencies for a comprehensive understanding of complex relationships in multi-dimensional data. This approach significantly enhances the accuracy and reliability of forecasts in multi-sensor environments.
\vspace{-1mm}
\item The LoRA-AMR method offers a trifecta of benefits: enhanced memory efficiency, consistent computational load, and reduced activation storage memory requirements, all without compromising inference latency. A larger LM, such as `Llama2-70B', is employed to generate textual descriptions of patterns and trends. This process facilitates a context-driven interpretation of MTS data. Fine-tuning the smaller `Llama2-7B' model using generated descriptions using the powerful LoRA-AMR technique not only enables efficient and effective customization of the smaller LM to the specific task but also leverages the advanced capabilities of larger LLMs to enhance the performance and versatility of smaller LMs.
\vspace{-1mm}
\item We integrate text-level embeddings obtained from fine-tuned smaller LMs and time series embeddings from traditional methods using a multi-head attention mechanism. This approach enables the capture of contextually relevant information from cross-domains, enhancing the analysis and understanding of MTS data.
\vspace{-2mm}
\end{itemize}

\vspace{-3mm}
\section{Problem Definition}
\vspace{-1mm}
Consider a dynamic system equipped with \( N \) sensors that gather sequential data over \( T \) time intervals on \( F \) input features, represented in a spatio-temporal matrix \( \mathbf{X} \in \mathbb{R}^{N \times T \times F} \). These features typically include key attributes such as traffic speed, flow, and density. Specifically, we define \( \mathbf{X}_{i} \in \mathbb{R}^{T \times F} \) as the historical data for the \( i \)-th sensor, encompassing all features over time, and \( \mathbf{X}^{t} \in \mathbb{R}^{N \times F} \) as the historical data for all sensors at time step \( t \), including all features. In our work, we predict traffic flow using only \( F=1 \) to ensure a fair and rigorous comparison with the baselines in traffic forecasting datasets. In the non-stationary time series data matrix \( \mathbf{X} \in \mathbb{R}^{N \times T} \), each row represents data from a sensor, while each column corresponds to data at a specific timestamp. We use subscripts and superscripts to denote data from a specific sensor and timestamp, respectively. For instance, \( \mathbf{X}_{i} = \mathbf{X}_{i,:} \) represents the time series data from sensor \( i \), and \( \mathbf{X}^{t} = \mathbf{X}_{:,t} \) represents the data across all sensors at timestamp \( t \). In the context of time series forecasting, we employ the sliding window technique to divide the historical dataset into overlapping, consecutive segments. This approach allows predictive models to adapt to changing patterns and complex dynamics, thereby effectively capturing both short-term and long-term trends, as well as seasonality and other dominant characteristics. We construct the samples \( \mathbf{X}^{t-W : t-1} \in \mathbb{R}^{N \times W \times F} \) using a sliding window of size \( W \). In traffic forecasting, our objective is to train a neural network, denoted as \( \boldsymbol{\Theta} \), to predict future data for the upcoming \( \nu \) steps, represented as \( \mathbf{S}^{t+1} = \mathbf{X}^{t: t + \nu - 1} \), based on historical observations \( \mathbf{S}^{t} = \mathbf{X}^{t-W: t-1} \). The process is illustrated as follows:

\vspace{-5mm}
\resizebox{0.99\linewidth}{!}{
\centering
\begin{minipage}{\linewidth}
\begin{equation}
    \mathbf{S}^{t} \xrightarrow{\boldsymbol{\Theta}} \mathbf{S}^{t+1} \nonumber
\end{equation}
\end{minipage}
}

\vspace{2mm}
The loss function to train the spatio-temporal encoder is mean absolute error(MAE) loss, defined as follows:

\vspace{0mm}
\resizebox{0.875\linewidth}{!}{
\begin{minipage}{\linewidth}
\begin{equation}
    \mathcal{L}(\mathbf{\Theta}) = \frac{1}{|\nu|} |\hat{\mathbf{S}}^{t+1} - \mathbf{S}^{t+1}| \nonumber
 \end{equation}
 \end{minipage}
}

where \( \hat{\mathbf{S}}^{t+1} \) represents the framework's predictions and \( \mathbf{S}^{t+1} \) is the ground truth. Specifically, we define \( \mathbf{S}^{t+1} = \mathbf{X}^{t: t + \nu - 1} \in \mathbb{R}^{N \times 1 \times F} \) for single-step forecasting, and \( \mathbf{S}^{t+1} = \mathbf{X}^{t: t + \nu - 1} \in \mathbb{R}^{N \times \nu \times F} \) for multi-step forecasting. Here, \( \nu \) represents the forecasting horizon.

\begin{figure}[ht!]
\centering
\resizebox{1.0\linewidth}{!}{ 
\hspace*{3mm}\includegraphics[keepaspectratio,height=4.5cm,trim=0.0cm 0.0cm 0cm 1.25cm,clip]{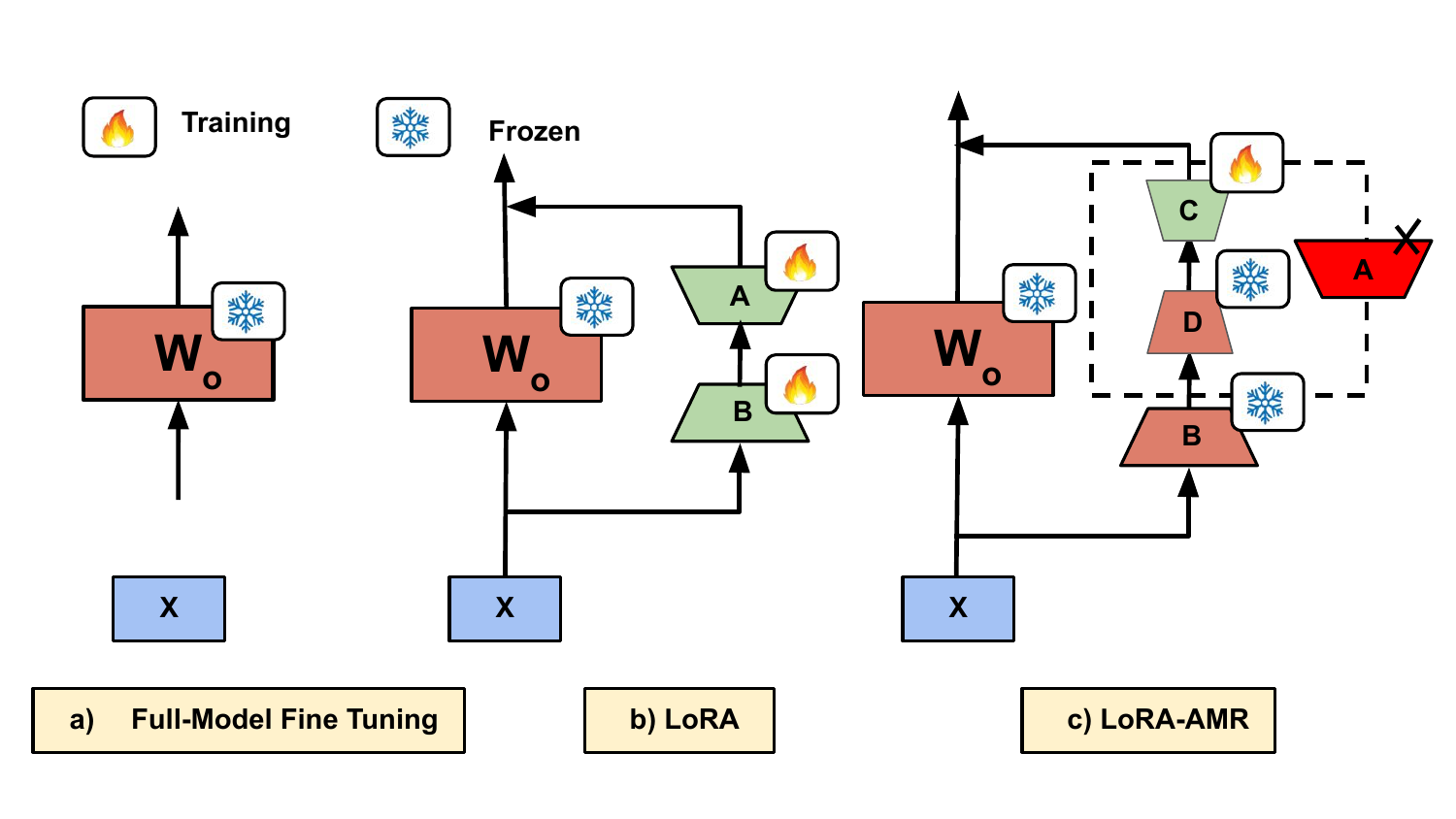} 
}
\vspace{-9mm}
\caption{The illustration of (a) full-model fine-tuning (FT), (b) LoRA, and (c) LoRA-AMR.}
\label{fig:figure2}
\vspace{-3mm}
\end{figure}

\vspace{-2mm}
\paragraph{LoRA-AMR:} LLMs have revolutionized natural language processing\cite{brown2020language,llama,gpt4,palm2}, and fine-tuning these large-scale models has proven to be highly effective in enhancing their performance across various tasks and in aligning with human intent\cite{roberta,flanv2}. However, fine-tuning LLMs with their full set of parameters can be extremely resource-intensive, especially for specialized tasks on consumer-grade hardware, due to memory constraints. To address this challenge, parameter-efficient fine-tuning (PEFT) methods have been introduced. These methods focus on updating only a small subset of the trainable parameters, such as adapter weights \cite{houlsby2019parameter} and prompt weights \cite{li2021prefix,lester2021power}. In particular, Low-Rank Adaptation (LoRA) \cite{lora}, is notable for fine-tuning pretrained LLMs to achieve performance comparable to full-parameter fine-tuning. It accomplishes this by updating only a small number of learnable pairs of low-rank adapters(weights) while keeping the base parameters static. This technique has been widely adopted in various applications \cite{qlora}, facilitating efficient task-specific adaptation. LoRA prevents catastrophic forgetting in general-purpose large-scale models during continual learning by enabling them to adapt to new tasks without overwriting the pretrained knowledge base. This allows them to retain their prior knowledge while effectively learning new information. In essence, LoRA represents a parameter-efficient adaptation technique that substantially enhances the capabilities of LLMs. It achieves this by integrating parallel low-rank adapters alongside the original weights of a linear layer, as depicted in Figure \ref{fig:figure2}(b). These adapters operate in conjunction with the frozen pre-trained weights($W_0$) of the linear layer. This approach significantly reduces memory usage while maintaining inference efficiency. It accomplishes this by keeping the primary weights static and updating only the lightweight, ancillary parameters---the LoRA adapters. LoRA aims to introduce additional trainable parameters ($\Delta W$) that capture task-specific information without altering the original pre-trained weights ($W_0$). To achieve this, LoRA constrains weight updates to a low-rank decomposition, expressed as $W_0 + \Delta W = W_0 + \alpha BA$, where $W_0$ represents the original pre-trained weight matrix with dimensions $\mathbb{R}^{d \times d}$. $\Delta W$ denotes the low-rank approximation added to the original weights during model adaptation (fine-tuning). This approximation is constructed as the product of two low-rank matrices, $B$ and $A$, both confined within a low-rank space. Here, $B$ is the projection-down weight matrix with dimensions $\mathbb{R}^{d \times r}$, $A$ is the projection-up weight matrix with dimensions $\mathbb{R}^{r \times d}$. The notation $r \ll d$ indicates that the rank of the decomposition is significantly smaller than $d$, leading to substantial memory savings. The hyper-parameter $\alpha$, typically valued at $\frac{1}{r}$, is a positive constant. The rank $r$ controls the trade-off between model capacity (how much task-specific information it can learn) and the complexity (number of parameters to train). During training, $W_0$ remains fixed, and only the low-rank weights $B$ and $A$ are updated. This approach reduces the memory overhead by decreasing the number of trainable parameters to update, the corresponding gradients to compute, and the optimizer state size that needs to be maintained. Compared to full-parameter fine-tuning, this method yields a parameter reduction ratio of $\frac{d}{2r}$, which is significant when the rank $r$ is much smaller than the dimension $d$. Furthermore, LoRA introduces no additional inference latency, as the product of $BA$ is added element-wise into $W_0$. However, LoRA faces challenges related to high memory usage during fine-tuning. This is due to the necessity of storing large input activations of $X$ throughout the forward-propagation phase for gradient computation of the weight matrix $A$ in the back-propagation phase. As a result, LoRA incurs high activation memory costs comparable to conventional full-parameter fine-tuning, potentially leading to a memory bottleneck. Current solutions include selectively applying LoRA to specific layers \cite{lora} or employing activation recomputation \cite{chen2016training} to mitigate this issue. However, these strategies might affect the fine-tuning performance and efficiency.

\vspace{-4mm}

\vspace{0mm}
In this work, we introduce LoRA-AMR, a novel approach designed to significantly reduce the activation memory footprint associated with LoRA without incurring additional computational costs. LoRA-AMR innovates by freezing the pre-trained weight $W_0$ as well as the projection-down weight $B$. Furthermore, it redefines the projection-up weight $A$ as the product of two low-rank matrices, $D$ and $C$. Specifically, $D$, the second projection-down weight, has dimensions $\mathbb{R}^{r \times \frac{r}{2}}$, and $C$, the second projection-up weight, has dimensions $\mathbb{R}^{\frac{r}{2} \times d}$. In our proposed method, $D$ remains static while only $C$ is updated during training. In short, by freezing $W_0$, $B$, and $D$, and updating only $C$, LoRA-AMR reduces the number of trainable parameters and minimizes the size of the input activations that must be stored during training for gradient computation during backward propagation, all with no additional inference latency. This modification results in a substantial decrease in the activation memory required, as it confines the input activation storage to the lower-dimensional output of matrix \(D\). Consequently, the only input that needs to be stored during the feed-forward pass is the much smaller transformed input \(DBX\), used to compute the gradient of \(C\) during the backward propagation. Here, \(X\) is first mapped through weights \(B\) and \(D\) to a lower dimension before being projected back up through weight \(C\), significantly reducing the activation memory demands. Figure \ref{fig:figure2} compares the full-model fine-tuning (FT), LoRA, and LoRA-AMR approaches. LoRA-AMR begins by initializing the low-rank parameters $B$ and $D$, which are randomly drawn from a normal distribution. It sets $C$ initially to zero. As a result, the adaptation weight matrix $\Delta W = BA = B(DC)$ remains zero at the start. This ensures that the outputs of the pretrained LLMs remain unchanged before fine-tuning begins. During fine-tuning, $B$ and $D$ are held fixed. This means that updates to the model weights through $\Delta W$ are confined to a subspace of reduced rank, specifically $r/2$, as determined by the initial column space of $D$. This approach strategically limits weight updates to this lower-dimensional space, preserving the integrity of the original LLMs predictions while the model undergoes task-specific adaptation.  LoRA-AMR is an efficient fine-tuning method for LLMs that reduces computational overhead and activation storage memory requirements without impacting inference latency. It simplifies fine-tuning by restricting updates to a low-rank weight, $C$, thus avoiding any increase in memory overhead. This is achieved by freezing weights and omitting gradient calculations for $B$ and $D$, effectively reducing expensive activation memory consumption. During inference, LoRA-AMR adds low-rank weights $\Delta W = BA = B(DC)$ element-wise directly into the pre-trained weight matrix $W_0$. This ensures no extra latency is introduced, thereby maintaining the efficiency of a fully fine-tuned model. In summary, LoRA-AMR presents a trifecta of benefits: a) It enhances memory efficiency by minimizing the number of trainable parameters by confining the weight adaptations to low-rank decompositions of matrices and shrinking the size of activations needed to be stored during training, b) It maintains a consistent computational load during the fine-tuning phase without adding to the latency during inference.

\vspace{-1mm}
\paragraph{Fine-Tuning small-scale LMs:} Llama 2 is an autoregressive, language-optimized transformer architecture designed for specific applications. It has been fine-tuned using supervised fine-tuning (SFT) and reinforcement learning with human feedback (RLHF) to align with human preferences for helpfulness and safety. Llama 2 introduces enhancements such as RMSNorm pre-normalization, PaLM-inspired SwiGLU activation functions, and rotary positional embeddings to improve performance and efficiency. The implementation of grouped-query attention extends its context length capability to 4096 tokens, significantly improving its ability to handle longer text sequences. The Llama2 model infact has 32 layers and 32 attention heads, with a hidden size (dimensionality) of 4096. It supports a maximum batch size of 32 and can output sequences up to 2048 tokens in length. We incorporate LoRA-AMR modules into every linear layer within the transformer blocks to augment the fine-tuning efficacy (Zhang et al., 2023), where these modules update only a small set of parameters added to the existing weights. This efficient updating, achieved by introducing low-rank matrices, captures the essential changes needed during fine-tuning, enabling the adaptation of a large model to the specialized task without extensive retraining and thus saving computational resources. Time series data, composed of numerical sequences, exhibits a structural similarity to language data. However, the elements of these sequences are numbers rather than words. Language models, recognized for their ability to produce coherent and contextually relevant text across a diverse range of topics, have the potential to excel at generating technical descriptions of trends and patterns in time series. Such descriptions would typically involve identifying notable upward or downward trends, analyzing seasonal patterns or cycles, pinpointing anomalies, and evaluating long-term trends. Despite their remarkable capabilities, language models face challenges with numerical data due to their tokenization methods. A prevalent technique, Byte-Pair Encoding (BPE), segments numbers into inconsistent segments based on their frequency in the training data, hindering the acquisition of fundamental mathematical operations. The LLaMA tokenizer alleviates this issue by tokenizing the numbers into individual digits, leading to a significant enhancement in mathematical performance. For example, the model can effectively tokenize the sequence `$123$' is composed of the individual digits `$1$', `$2$', `$3$'. The Llama 2 architecture employs a 16-bit format for its model weights, implying that each weight occupies 2 bytes of memory. This translates to a memory cost of $2m$, where $m$ represents the number of model parameters. For instance, decoder-based NLG models like LLaMA-7B incur a memory cost of 14 GB for model weights. However, this 14 GB requirement exceeds the memory capacity of a single consumer GPU, such as those in the Nvidia GeForce series, which typically have a 12 GB GPU memory. In LoRA-AMR, the pretrained model weights are frozen during fine-tuning, allowing them to be quantized into lower bit widths without compromising fine-tuning performance. This quantization can significantly reduce both computational cost and memory footprint. For instance, combining 4-bit quantization method with LoRA-AMR can reduce the model weight memory by up to 4 times. This reduction in bit-width leads to substantial savings in memory and computational resources, thereby enhancing the model's efficiency during both training and inference. Quantization is the process of converting a real number into a fixed-point representation. In 4-bit quantization, a real number is converted into an 4-bit fixed-point representation. This technique can reduce the memory footprint of a large model by quantizing its weights, making LoRA models more accessible for fine-tuning with limited hardware resources for task-specific customization. One advantage of LoRA-AMR is that it can quantize LLMs to 4 bits without significant accuracy loss, as it effectively compensates through the low-rank adapters for the typical accuracy reduction associated with 4-bit quantization. However, only the original, pre-trained weights that are frozen are quantized. The low-rank adapters that are updated are in the 16-bit format. During inference, the frozen 4-bit parameters are dequantized to 16-bit for backpropagation. We use zero-shot prompting to instruct open-source LLMs, such as Llama2-70B, to generate technical descriptions of time series trend analysis. Llama2-70B-assisted time series trend analysis data generation is a method for creating high-quality training data for fine-tuning smaller Llama2-7B models. Llama2-70B-assisted time series trend analysis data generation offers several advantages over traditional methods, such as manual analysis by humans. It can process and interpret complex datasets that might be challenging for humans, especially those involving multivariate relationships or nonlinear trends. We employ supervised fine-tuning of a smaller Llama2-7B model, utilizing the LoRA-AMR technique and 8-bit quantization, based on provided input-output pairs of time series data and corresponding textual descriptions, minimizing cross-entropy loss. This approach proves particularly effective for sequence generation tasks, such as generating technical descriptions for unseen time series data. Supervised fine-tuning emerges as a powerful technique, enabling the adaptation of smaller Llama2-7B models to time series trend analysis, even if such tasks were not explicitly considered during the pre-training of the smaller language model. After fine-tuning a small-scale Llama2-7B model using the technical descriptions generated by Llama2-70B for domain-specific customization. We input text sequences from the Llama2-70B model, denoted as $\mathcal{S}_{e} \in \mathbb{R}^{N \times W \times m}$, into the smaller Llama2-7B model, referred to as the $\textrm{LM}_{\textrm{expl}}$ model, to compute expressive token embeddings that capture contextual information and semantic relationships between words or phrases:

\vspace{-1mm}
\resizebox{0.965\linewidth}{!}{
\hspace{0mm}\begin{minipage}{\linewidth}
\begin{equation}
H_{\textrm{expl}} = \textrm{LM}_\textrm{expl}(\mathcal{S}_{e}) \nonumber
\end{equation}
\end{minipage}
}

\vspace{1mm}
where $H_{\textrm{expl}} \in \mathbb{R}^{N \times W \times m \times d}$ denotes the context-aware token embeddings, with $m$ representing the number of tokens in the input sequence $\mathcal{S}_{e}$, $d$ being the dimensionality of token embedding, $N$ denoting the sensors, and $W$ the window size. We apply a softmax attention pooling mechanism to these contextualized embeddings to compute text-level embeddings for each sensor at each historical window step, which captures the rich domain-specific knowledge embedded within the generated textual descriptions:

\vspace{-1mm}
\resizebox{0.965\linewidth}{!}{
\hspace{0mm}\begin{minipage}{\linewidth}
\begin{equation}
\alpha_i = \mbox{softmax}(q_i); \hspace{2mm} q_i = \mathbf{u}^T H_{\textrm{expl}} \nonumber
\end{equation}
\end{minipage}
}

\vspace{0mm}
\resizebox{0.935\linewidth}{!}{
\hspace{0mm}\begin{minipage}{\linewidth}
\begin{equation}
H_{\text{text}} = \sum_{i=0}^m{\alpha_i H^{(i)}_{\textrm{expl}}} \nonumber
\end{equation}
\end{minipage}
}

\vspace{0mm}
where $\mathbf{u}$ is a trainable parameter and updated based on the downstream forecasting task. $\alpha$ is the attention coefficient. The text-level embedding $H_{\text{text}} \in \mathbb{R}^{N \times W \times d}$ encapsulates domain-specific knowledge from the foundational Llama2-70B on MTS data trend analysis, enhancing explainability by unpacking the black-box nature of Llama2-70B. This approach also improves the Llama2-7B model's ability to interpret and analyze time series data with high precision, leveraging the nuanced insights provided by the advanced analysis capabilities of Llama2-70B. 

\vspace{-2mm}
\paragraph{Dynamic Prompting Mechanism Design:} Instruction-Following LLMs leverage large-scale models' knowledge using prompt engineering to generate user query responses. This eliminates fine-tuning or retraining for task-specific adaptation. However, fixed prompts can lead to suboptimal performance, highlighting the potential for refined prompt design to enhance model performance across applications. Traditional forecasting models face a similar challenge in their limited adaptability for diverse forecasting scenarios. Their inability to process dynamic, non-stationary data hinders their predictive accuracy and applicability, often arising from the use of fixed historical data window sizes. Furthermore, their limited ability to handle non-stationary distributions, capture long-range dependencies, and predict complex relationships among time series variables contributes to this limitation. Identifying these latent relationships is crucial for uncovering non-obvious patterns, which are vital to effective time series modeling. Analogous to language models, dynamic prompting in time series forecasting addresses the limitations of traditional methods, enabling more adaptable and accurate predictions in response to ever-evolving MTS data patterns.
We introduce a dynamic prompting mechanism that enables traditional forecasting methods to access relevant past knowledge and apply it to new, similar time series data. This enhancement improves the methods ability to adapt to different types of time series patterns or trends, thereby increasing their efficiency and accuracy. Our work employs a predefined window length, which may not be optimal for capturing the evolution of temporal patterns and interrelated variable dynamics over time, given the varying durations of these patterns. An excessively large window length can obscure short-range patterns within the data, while an excessively small window may fail to capture long-range patterns. To address these limitations, we employ prompts specifically designed to capture various time series characteristics and utilize retrieval-based prompt selection for temporal reasoning and knowledge sharing across the time series data. This approach facilitates the identification and application of learned patterns for forecasting tasks. To address the dynamic nature of real-world MTS data, which is often characterized by distributional shifts, our approach introduces a shared pool of prompts, each uniquely identified by distinct key-value pairs. This prompt pool consists of key-value pairs, where each key is a vector in the embedding space and each value is a corresponding prompt representation. This design aims to enable the framework to effectively draw upon relevant past knowledge, particularly in scenarios where the input time series data resembles or is similar to past data. This allows the framework to access a corresponding set of prompts from this collective pool, enabling the model to selectively identify and use the most fitting and appropriate prompts for each specific instance of time series data.  This approach enhances modeling efficiency and predictive accuracy by empowering the framework to identify and employ patterns across the historical data leveraging a shared prompt pool. These prompts encapsulate various time series characteristics, such as temporal dependencies, periodic trends, and seasonality pertinent to diverse time periods, facilitating the model's adaptation to new and varied data.  The shared pool of prompts stored as key-value pairs is defined as follows:

\vspace{-1mm}
\resizebox{0.95\linewidth}{!}{
\hspace{0mm}\begin{minipage}{\linewidth}
\begin{equation}
    \mathbf{V}_\mathbf{K}=\left\{\left(\boldsymbol{k}_1, V_1\right),\left(\boldsymbol{k}_2, V_2\right), \cdots,\left(\boldsymbol{k}_M, V_M\right)\right\} \nonumber
\end{equation}
\end{minipage}
}

Where $M$ is the predefined length of the prompt pool,  $V_m \in \mathbb{R}^{ W \times d}$ represents a single prompt with a token length of $W$ and an embedding size of $d$ in the set $\mathbf{V} = \left\{ V_m \right\}_{m=1}^M$, and $\boldsymbol{k}_m \in \mathbb{R}^{d}$ is the key for each prompt in the set $\mathbf{K} = \left\{ \boldsymbol{k}_m \right\}_{m=1}^M$. We project the initial time series data of each sensor $\mathbf{S}^{t}_{i} \in \mathbb{R}^{W \times 1}$ through a shared linear layer to transform it into a $d$-dimensional space $\mathbf{S}^{t}_{i} \in \mathbb{R}^{W \times d}$. Each prompt in this key-value pair method is designed to encapsulate specific domain knowledge relevant to the forecasting task, enabling the model to utilize pre-existing insights to enhance its predictions. The time series data can attend to multiple prompts, collectively encoding knowledge relevant to the forecasting task, encompassing various time series characteristics. By learning to associate different prompts with diverse types of time series behavior, the framework effectively handles a wide spectrum of patterns, even those not encountered during training, leading to a deeper understanding and more accurate predictions of complex behaviors. We employ an additive attention mechanism to calculate the relevance of a given prompt (key-value pair) to the current input time series query $\mathbf{S}^{t}_{i}$, enabling efficient retrieval of the most pertinent prompts to facilitate dynamic adaptation to input data. Given a query $\mathbf{S}^{t}_{i} \in \mathbb{R}^{W \times d}$ and a key $\boldsymbol{k}_m \in \mathbb{R}^{d}$, the score-matching function is denoted as follows:

\vspace{0mm}
\resizebox{0.95\linewidth}{!}{
\hspace{0mm}\begin{minipage}{\linewidth}
\begin{equation}
    a(\mathbf{S}^{t}_{i}, \boldsymbol{k}_{m}) = \mathbf{W}_v^\top \textrm{tanh}(\mathbf{W}_q \mathbf{S}^{t}_{i} + \mathbf{W}_k \boldsymbol{k}_{m}) \in \mathbb{R} \nonumber
\end{equation}
\end{minipage}
}

\vspace{0.5mm}
where $\mathbf{W}_q \in \mathbb{R}^{W \times d}$, $\mathbf{W}_k \in \mathbb{R}^{d \times d}$, $\mathbf{W}_v \in \mathbb{R}^{d}$ and  $a: \mathbb{R}^{W \times d} \times \mathbb{R}^{d} \rightarrow \mathbb{R}$. This function identifies the most similar prompt keys and selects the corresponding prompt values. Through this score-matching function, the framework learns to optimize its predictions based on the retrieved prompts, resulting in overall improved forecasting performance. We then retrieve the top-\(\mathbf{K}\) corresponding prompt values for the input time series query \(\mathbf{S}^{t}_{i}\) and concatenate them to obtain the time series embedding for each sensor \(i\) as follows,

\vspace{-0.5mm}
\resizebox{0.95\linewidth}{!}{
\hspace{0mm}\begin{minipage}{\linewidth}
\begin{equation}
    \mathbf{\tilde{S}}^{t}_{i} = \left[V_{1} ; \cdots ; V_{\mathbf{K}} ; \mathbf{S}^{t}_{i}\right]\mathbf{W}_{o}, \quad 1 \leq \mathbf{K} \leq M \nonumber
\end{equation}
\end{minipage}
}

where  the linear layer (or fully connected layer) $\mathbf{W}_q \in \mathbb{R}^{((K+1) \times d) \times d)}$ and $\mathbf{\tilde{S}}^{t}_{i} \in \mathbb{R}^{W \times d}$. In summary, the framework employs a static prompt pool and a selection-based prompt method to efficiently train the framework for forecasting tasks. This approach facilitates the reuse of learnable continuous vector representations that encapsulate temporal knowledge, enabling the framework to adapt to evolving non-stationary time series data distributions and maintain forecasting accuracy as time series data progresses. Building upon the foundation established by our dynamic prompt pool, we now turn our attention to learning the temporal dynamics within MTS data. This progression from employing a diverse prompt pool to exploring the temporal intricacies enables our model to not only recognize but also profoundly comprehend the evolving relationships in time series data.

\vspace{-2.5mm}
\paragraph{Modeling Intra-series dependencies:} In this section, we aim to model the temporal dynamics of MTS data. Our goal is to accurately learn the intra-series dependencies within a single time series variable for improved pointwise forecasts. In this context, given the contextualized time series embedding, denoted as $\mathbf{\tilde{S}}^{t} \in \mathbb{R}^{N \times W \times d}$, we utilize the Grouped-query Multi-head Attention mechanism (GQ-MHA) to capture the non-linear, time-evolving dependencies underlying MTS data. This involves projecting the time series embedding for each group $g$ and each of the $N$ sensors. The time series embedding $\mathbf{\tilde{S}}_n^{t} \in \mathbb{R}^{W \times d}$ for sensor $n$ is projected to form shared keys $K_g$, shared values $V_g$, and a unique query projection $Q_{g,h}$ for each head in the group as follows:

\vspace{-3mm}
\resizebox{0.925\linewidth}{!}{
\begin{minipage}{\linewidth}
\begin{align*}
   K_g^n &= \mathbf{\tilde{S}}_n^{t} W_{K_g}, \quad n = 1, \ldots, N \\
   V_g^n &= \mathbf{\tilde{S}}_n^{t} W_{V_g}, \quad n = 1, \ldots, N \\
   Q_{g,h}^n &= \mathbf{\tilde{S}}_n^{t} W_{Q_{g,h}}, \quad n = 1, \ldots, N.
\end{align*}
 \end{minipage}
}

\vspace{1mm}
Here, the weight matrices $W_{K_g}$, $W_{V_g}$, and $W_{Q_{g,h}}$ have dimensions $\mathbb{R}^{d \times d}$. Consequently, the dimensions of $K_g^n$, $V_g^n$, and $Q_{g,h}^n$ for each sensor $n$ are $\mathbb{R}^{W \times d}$, respectively. We then calculate the temporal attention transformed embeddings. For each sensor $n$ and each head $h$ within the group $g$, the transformed time series embeddings are computed using the scaled dot-product attention mechanism as follows:

\vspace{-3mm}
\resizebox{0.925\linewidth}{!}{
\begin{minipage}{\linewidth}
\begin{equation}
\text{Attention}(Q_{g,h}^n, K_g^n, V_g^n) = \text{softmax}\left( \frac{Q_{g,h}^n (K_g^n)^T}{\sqrt{d_k}} \right) V_g^n \nonumber
\end{equation}
 \end{minipage}
}  

where $d_k = \frac{d}{H}$ is a scaling factor and $H$ is the total number of heads. We perform aggregation across heads for each sensor in a group $g$. For aggregating across multiple groups, assuming there are $G$ groups, we average the outputs from each group for each sensor as follows:

\vspace{-3mm}
\resizebox{0.925\linewidth}{!}{
\begin{minipage}{\linewidth}
\begin{equation}
\mathbf{\tilde{S}}_{n}^{t} = \frac{1}{G} \sum_{g=1}^G(\text{Concat}(\text{Attention}_1, \ldots, \text{Attention}_H) W_o) \nonumber
\end{equation}
 \end{minipage}
}

Here, $W_o \in \mathbb{R}^{Hd \times d}$. These steps of head-level and group-level aggregation are crucial for synthesizing a concise representation of the time series data. In conclusion, the integration of a flexible retrieval-based prompt pool with an advanced attention mechanism for learning intra-series dependencies represents a comprehensive approach to time series modeling. Intra-series modeling focuses on understanding the temporal dynamics within a single univariate variable, while inter-series modeling broadens this scope to include the dependencies among different variables. Together, they aim to capture the full spectrum of dependencies in spatio-temporal MTS data. The next section will discuss the inter-series modeling approach and its contribution to the development of an integrated analytical framework for enhancing our understanding of the multi-dimensional nature of data.

\vspace{-5mm}
\paragraph{Modeling Inter-series dependencies:} In this section, we aim to model and learn the inter-dependencies among variables underlying the spatio-temporal MTS data to provide accurate pointwise forecasts. In this context, given the contextualized time series embedding, denoted as $\mathbf{\tilde{S}}^{t} \in \mathbb{R}^{N \times W \times d}$, the goal is to utilize the Grouped-query Multi-head Attention mechanism (GQ-MHA) to capture the spatial dependencies among different variables at each time step. This involves projecting the time series embedding for each window step $w$ and each group $g$. The time series embedding $\mathbf{\tilde{S}}_w^{t} \in \mathbb{R}^{N \times d}$ for window step $w$ is projected into shared keys $K_g$, shared values $V_g$ for each group $g$, and a unique query projection $Q_{g,h}$ for each head $h$ in the group $g$ as follows,

\vspace{-2.5mm}
\resizebox{0.925\linewidth}{!}{
\begin{minipage}{\linewidth}
\begin{align*}
   K_g^w &= \mathbf{\tilde{S}}_w^{t} W_{K_g}, \quad w = 1, \ldots, W \\
   V_g^w &= \mathbf{\tilde{S}}_w^{t} W_{V_g}, \quad w = 1, \ldots, W \\
   Q_{g,h}^w &= \mathbf{\tilde{S}}_w^{t} W_{Q_{g,h}}, \quad w = 1, \ldots, W.
\end{align*}
\end{minipage}
}

\vspace{1mm}
Here, the weight matrices $W_{K_g}$, $W_{V_g}$, and $W_{Q_{g,h}}$ have dimensions $\mathbb{R}^{d \times d}$. Consequently, the dimensions of $K_g^w$, $V_g^w$, and $Q_{g,h}^w$ for each window step $w$ are $\mathbb{R}^{N \times d}$, respectively. The GQ-MHA mechanism calculates the attention scores at each window step $w$. This mechanism is pivotal for understanding the inter-dependencies among different variables in the spatio-temporal MTS data. Specifically, for each window step $w$ and each head $h$ within the group $g$, the spatial attention transformed embeddings are computed using the scaled dot-product attention mechanism as follows:

\vspace{-2mm}
\resizebox{0.925\linewidth}{!}{
\begin{minipage}{\linewidth}
\begin{equation}
\text{Attention}(Q_{g,h}^w, K_g^w, V_g^w) = \text{softmax}\left( \frac{Q_{g,h}^w (K_g^w)^T}{\sqrt{d_k}} \right) V_g^w \nonumber
\end{equation}
\end{minipage}
}

Here, $d_k = \frac{d}{H}$ is the scaling factor, and $H$ denotes the total number of heads in the attention mechanism. After computing the attention scores, the framework aggregates these scores across all heads and groups to synthesize a comprehensive representation. The aggregation is performed as follows:

\vspace{-3mm}
\resizebox{0.925\linewidth}{!}{
\begin{minipage}{\linewidth}
\begin{equation}
\mathbf{\tilde{S}}_{w}^{t} = \frac{1}{G} \sum_{g=1}^G(\text{Concat}(\text{Attention}_1, \ldots, \text{Attention}_H) W_O) \nonumber
\end{equation}
\end{minipage}
}

where $W_O \in \mathbb{R}^{Hd \times d}$. In conclusion, the utilization of the GQ-MHA mechanism for modeling both intra-series and inter-series dependencies offers a robust framework for forecasting in spatio-temporal MTS data. By capturing the complex, non-linear relationships both within and across different series, this approach significantly enhances the accuracy and reliability of forecasts in multi-sensor environments.

\vspace{-2mm}
\paragraph{Output Layer:} We employ the multi-head attention mechanism (MHA)\cite{vaswani2017attention} to integrate the text-level embeddings \( H_{\text{text}} \) with the spatio-temporal attention transformed time series embeddings \( \mathbf{\tilde{S}}^{t} \). This integration enables the capture of contextually relevant information and achieves semantic alignment across different cross-domain embeddings, resulting in the framework predictions \( \hat{\mathbf{S}}^{t+1} \). Thus, by focusing on and aligning high-level textual descriptions (text-level embeddings) with expressive time series embeddings, we ensure a comprehensive understanding and analysis of MTS data from both perspectives.

\vspace{-2mm}
\begin{table}[tbhp]
\center
\setlength{\tabcolsep}{0.25em} 
\renewcommand\arraystretch{1.325} 
\centering
\resizebox{0.50\textwidth}{!}{
\hspace{-5mm}\begin{tabular}{c|c|c|c|c|c}
\hline
\textbf{Dataset} & \textbf{Nodes} & \textbf{Timesteps} & \textbf{Time-Range} & \multicolumn{1}{l|}{\textbf{Data Split}} & \multicolumn{1}{l}{\textbf{Granularity}} \\ \hline
PeMSD3 & 358 & 26,208 & 09/2018 - 11/2018 & \multirow{5}{*}{6 / 2 / 2} & \multirow{7}{*}{\rotatebox[origin=c]{270}{5 mins}} \\
PeMSD4 & 307 & 16,992 & 01/2018 - 02/2018 &  &  \\
PeMSD7 & 883 & 28,224 & 05/2017 - 08/2017 &  &  \\
PeMSD8 & 170 & 17,856 & 07/2016 - 08/2016 &  &  \\
PeMSD7(M) & 228 & 12,672 & 05/2012 - 06/2012 &  &  \\ \cline{1-5}
METR-LA & 207 & 34,272 & 03/2012 - 06/2012 & \multirow{2}{*}{7 / 1 / 2} &  \\
PEMS-BAY & 325 & 52,116 & 01/2017 - 05/2017 &  &  \\ \hline
\end{tabular}
}
\vspace{-2.5mm}
\caption{The table provides a comprehensive overview of traffic datasets used in the MTSF task, detailing the timestamps, time range, data split, and granularity.}
\label{tab:summarydatasets}
\vspace{-5mm}
\end{table} 

\vspace{-2mm}
\section{DATASETS}
\vspace{-1mm}
Our study evaluates the proposed framework, \textbf{LLM-TS Net}, and its variant for estimating the uncertainty of future estimates, \textbf{w/Unc-LLM-TS Net}, on large-scale spatial-temporal traffic datasets (PeMSD3, PeMSD4, PeMSD7, PeMSD7(M), PeMSD8) obtained from the Caltrans Performance Measurement System (PeMS, \cite{chen2001freeway}). PeMS is a crucial system for providing real-time and historical traffic data that is essential for traffic management, monitoring, and analysis on California's freeways. The primary purpose of this system is to collect and assess real-time traffic information from an extensive network of detectors. Table \ref{tab:summarydatasets} provides details about the benchmark datasets. To maintain consistency with previous research, we converted the 30-second interval data into 5-minute averages, adhering to the method presented by \cite{choi2022graph}. Additionally, we employed publicly available traffic flow datasets (METR-LA and PEMS-BAY) obtained from \cite{LiYS018}, converting them into 5-minute interval averages. This consistent data format enabled us to demonstrate the effectiveness of our methodology in analyzing and modeling complex spatio-temporal MTS data, surpassing the performance of existing methods. 

\vspace{-4mm}
\section{EXPERIMENTAL RESULTS}
\vspace{-1mm}
Table \ref{tab:results1} presents a detailed comparison between the proposed models, namely \textbf{LLM-TS Net} and \textbf{w/Unc-LLM-TS Net}, and various baseline models in the MTSF task. This evaluation encompasses five benchmark datasets: PeMSD3, PeMSD4, PeMSD7, PeMSD7M, and PeMSD8. The performance of the proposed models was assessed by measuring forecast errors for a recognized benchmarking task that involves using historical data from 12 time steps prior to predicting estimates 12 time steps into the future. Our assessment employed a multi-metric approach in multi-horizon prediction tasks to comprehensively evaluate the models performance compared to baseline models. To ensure a thorough and robust analysis, we utilized various performance metrics, including mean absolute error (MAE), root mean squared error (RMSE), and mean absolute percentage error (MAPE), to gauge the effectiveness of the models. In this study, we report the results of baseline models from \cite{choi2022graph}. Our experimental findings demonstrate that the proposed models, \textbf{LLM-TS Net} and \textbf{w/Unc-LLM-TS Net}, consistently outperform the baseline models in terms of performance, achieving lower forecast errors across a variety of benchmark datasets. The empirical results highlight the effectiveness of this proposed neural forecasting architecture in capturing the complex and nonlinear spatio-temporal dynamics in MTS data, resulting in improved forecasts. The \textbf{w/Unc-LLM-TS Net} model, which integrates \textbf{LLM-TS Net} with local uncertainty estimation, not only provides pointwise forecasts but also predicts time-varying uncertainty estimates in its predictions. While it exhibits slightly lower performance compared to the \textbf{LLM-TS Net} model, it still surpasses several robust baselines in existing literature, as evidenced by its reduced prediction error. Additional experimental results, experimental setup, uncertainty estimation are discussed in detail in the technical appendix.

\vspace{-3mm}
\section{CONCLUSION}
\vspace{-1mm}
Our framework combines fine-tuning general-purpose LLMs with time series representation learning techniques aiming to deeply understand the spatial-temporal dynamics in MTS data, leading to precise multi-horizon forecasting. Experimental validation on real-world datasets confirms the effectiveness of our approach, as evidenced by enhanced multi-horizon forecasts and reliable uncertainty estimations.

\vspace{-3mm}
\bibliography{aaai24}

\vspace{-3mm}
\section{Technical Appendix}
\vspace{-1mm}
Table \ref{tab:results2} presents a comparative analysis in terms of the percentage increase in the performance of the proposed traffic prediction models, \textbf{LLM-TS Net} and \textbf{w/Unc-LLM-TS Net}, compared to the baselines. The analysis covers five benchmark PeMS datasets (PeMSD3, PeMSD4, PeMSD7, PeMSD8, and PeMSD7(M)) and employs three key performance metrics: Mean Absolute Error (MAE), Root Mean Square Error (RMSE), and Mean Absolute Percentage Error (MAPE). The results reveal a consistent trend of performance improvement for both models over the next-best baseline models, with \textbf{w/Unc-LLM-TS Net} slightly outperforming \textbf{LLM-TS Net} in most scenarios. This trend holds across all datasets, except for a notable exception in the PeMSD7 dataset under the MAPE metric, where \textbf{w/Unc-LLM-TS Net} exhibits a marginal decrease in performance. Particularly, \textbf{w/Unc-LLM-TS Net} excels in reducing MAE and RMSE across all datasets, indicating its robustness and reliability in traffic prediction. These results underscore the effectiveness of these models in traffic prediction tasks and highlight the significant advancements in model accuracy and reliability, positioning these models as strong contenders in the field of traffic prediction and analysis.

\vspace{-1mm}
\subsection{Additional Datasets and Experimental Results}
\begin{figure*}[h]
    \centering
    \subfloat[Predictions for forecast horizon@3]{
        \includegraphics[width=0.475\linewidth]{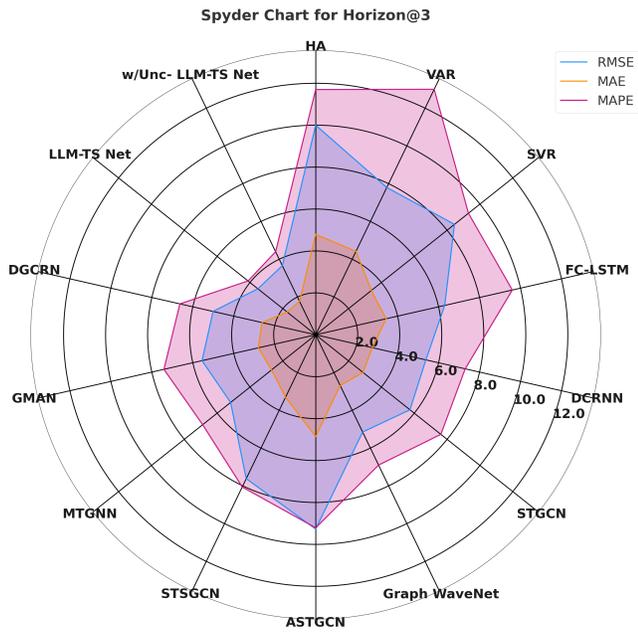}
        \label{fig:sub1}
    }
    \hfill
    \subfloat[Predictions for forecast horizon@6]{
        \includegraphics[width=0.475\linewidth]{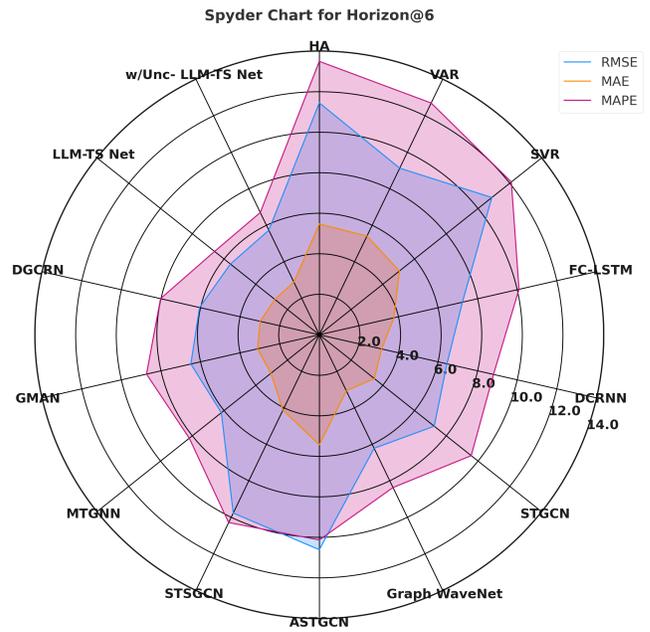}
        \label{fig:sub2}
    }
    
    \vspace{1em} 

    \subfloat[Predictions for forecast horizon@12]{
        \includegraphics[width=0.475\linewidth]{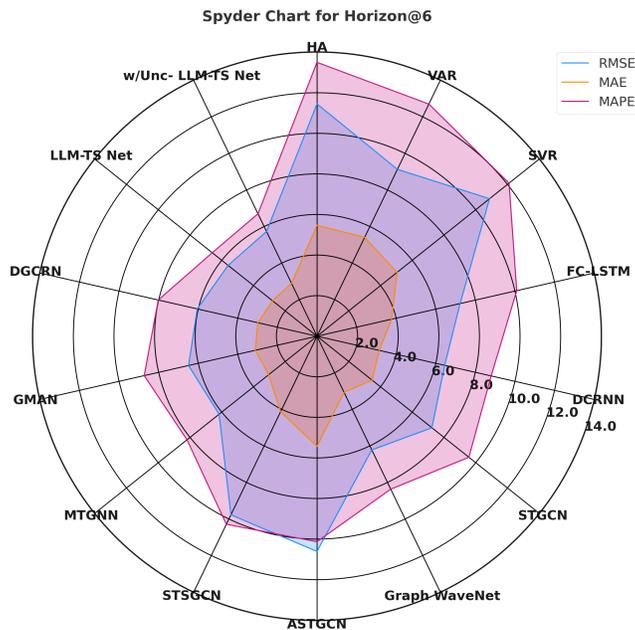}
        \label{fig:sub3}
    }
    \caption{The figure shows the comparative performance of different forecasting models on the \textbf{METR-LA} dataset. It showcases the accuracy and precision of each model in predicting traffic flow trends, emphasizing their respective strengths and limitations.}
    \label{fig:main1}
\end{figure*}

\begin{figure*}[h]
    \centering
    \subfloat[Predictions for forecast horizon@3]{
        \includegraphics[width=0.475\linewidth]{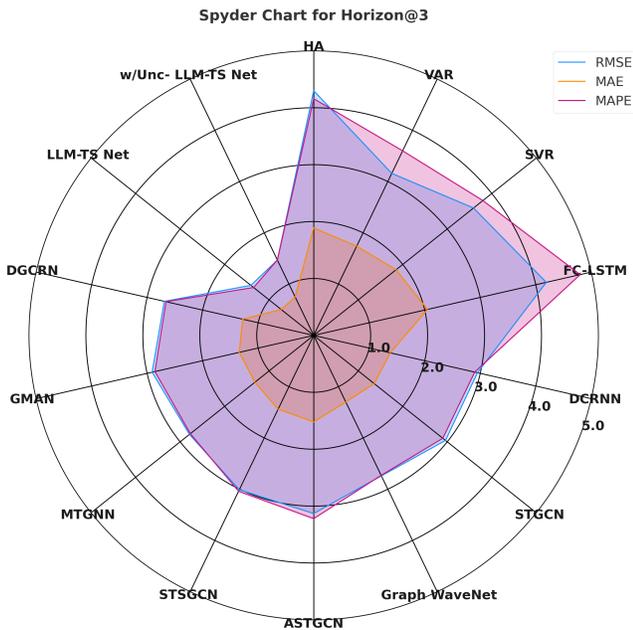}
        \label{fig:sub1}
    }
    \hfill
    \subfloat[Predictions for forecast horizon@6]{
        \includegraphics[width=0.475\linewidth]{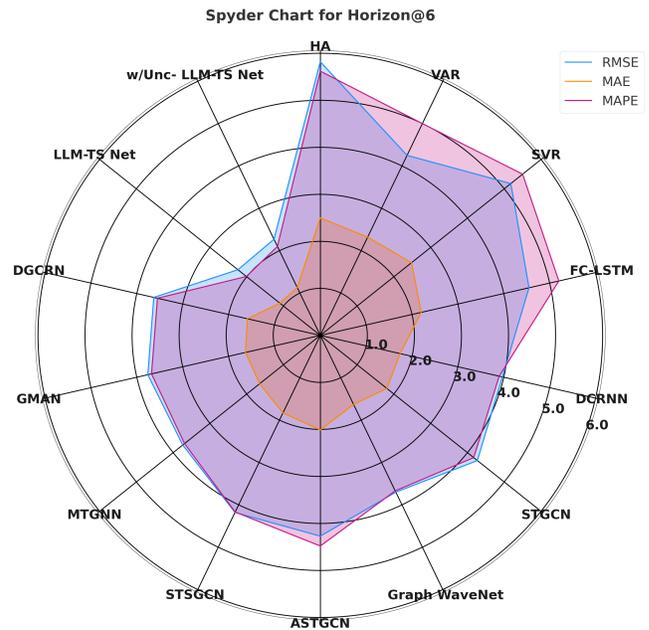}
        \label{fig:sub2}
    }
    
    \vspace{1em} 

    \subfloat[Predictions for forecast horizon@12]{
        \includegraphics[width=0.475\linewidth]{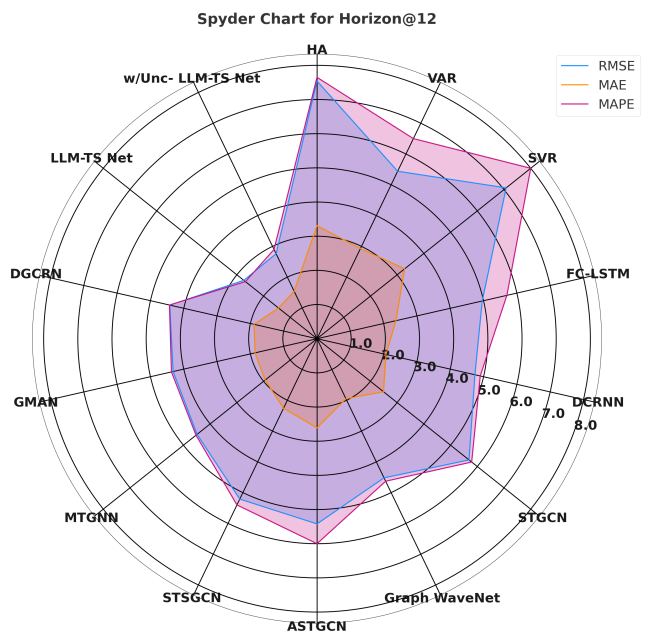}
        \label{fig:sub3}
    }
    \caption{The figure illustrates the performance comparison of various models on the \textbf{PEMS-BAY} dataset. It highlights the forecast accuracy and error margins for each model, providing insights into their relative effectiveness in predicting traffic flow patterns.}
    \label{fig:main2}
\end{figure*}

\vspace{-1mm}
Figures \ref{fig:main1} and \ref{fig:main2} illustrate the performance comparison of various models, including \textbf{LLM-TS Net}, \textbf{w/Unc-LLM-TS Net}, and several baseline models, on two datasets (METR-LA and PEMS-BAY) for the MTSF task. These models were evaluated using key metrics such as MAE, RMSE, and MAPE, with their forecast accuracy measured at 3-, 6-, and 12-steps ahead forecast horizons. Lower error in these forecasts indicates better model performance, highlighting their effectiveness in MTSF tasks. The baseline model results were reported from a prior study\cite{jiang2021dl}. Our experimental study revealed that our proposed model, \textbf{LLM-TS Net}, surpassed the baseline models in all the mentioned evaluation metrics across various forecasting horizons. In detail, the \textbf{LLM-TS Net} model demonstrated consistently higher accuracy and lower error rates compared to the baseline models. Additionally, Table \ref{tab:results3} presents a comparative analysis, showing the percentage performance improvement of the \textbf{LLM-TS Net} and \textbf{w/Unc-LLM-TS Net} models over the next-best SOTA baseline across the benchmark datasets. The comparative performance of the \textbf{LLM-TS Net} and \textbf{w/Unc-LLM-TS Net} models on the METR-LA and PEMS-BAY datasets reveals significant improvements in forecasting accuracy over baseline models by non-trivial margins. The \textbf{LLM-TS Net} model, in particular, exhibits notable enhancements across all metrics, with the highest increases observed in the shorter prediction intervals (MAPE@3, MAE@3, RMSE@3). This is especially evident in the PEMS-BAY dataset, where improvements like 49.25$\%$  in MAPE@3 and 47.58$\%$  in RMSE@3 are remarkable. The \textbf{w/Unc-LLM-TS Net} model also demonstrates substantial improvements, though slightly lower than its counterpart, following a similar trend of higher performance in shorter prediction intervals and a gradual decrease in improvement at longer intervals. Across both datasets, the \textbf{LLM-TS Net} generally outperforms the \textbf{w/Unc-LLM-TS Net}, and both models tend to perform better on the PEMS-BAY dataset compared to METR-LA. These trends, highlighting the superiority of these models in time series forecasting, are crucial for understanding their predictive capabilities and are significant.  
 
\vspace{-1mm}
\subsection{Ablation Study}
\vspace{-1mm}
 The proposed LLM-TS Net model introduces a unified framework that integrates various components to enhance the accuracy and reliability of forecasting in spatio-temporal MTS data. The overall framework architecture is illustrated in Figure \ref{fig:OverallArch}. (i) The first key component leverages both large-scale language models (LLMs) like Llama2-70B for generating textual descriptions of time series trends and smaller LLMs such as Llama2-70B for fine-tuning on generated descriptions for task-specific adaptation. This allows for the computation of text-level embeddings, capturing the insights contained within the descriptions. This approach takes advantage of the advanced reasoning and inference capabilities of large-scale LLMs, while the smaller LLMs provide task-specific fine-tuning that is both cost-effective and efficient. It utilizes the LoRA-AMR technique for fine-tuning smaller LMs, an efficient fine-tuning method that reduces computational overhead and activation storage memory requirements without affecting inference latency. This method enhances memory efficiency by limiting updates to low-rank weight adapters, avoiding expensive activation memory consumption. (ii) The next component for time-series representation learning involves the following methods: (a) Dynamic Prompting Mechanism: This mechanism is tailored to utilize learned historical patterns in time series data and adapt to emerging trends. It utilizes a pool of prompts that facilitate the transfer of pertinent historical knowledge for understanding and adapting to new data patterns. This involves a retrieval-based prompt selection approach, using prompts that each represent distinct time series characteristics to guide the model in forecasting tasks. The prompts are selected based on their relevance to the input data. (b) Modeling Intra-Series Dependencies: This method focuses on understanding and capturing the internal dynamics within each individual time series. It involves analyzing how each variable in the multivariate time series data evolves over time, which is crucial for accurate and robust forecasting in complex, real-world scenarios. (c) Modeling Inter-Series Dependencies: This aspect of the framework focuses on understanding the relationships between different variables in the MTS data. It analyzes the spatial dependencies among different variables at each time step. By doing so, it can accurately capture the complex, interrelated dynamics present in spatio-temporal MTS data. In summary, given a time series forecasting model utilizing a dynamic prompt-based mechanism with a sufficiently large and diverse prompt pool, the model can approximate any continuous time series forecasting function to an arbitrary degree of accuracy, under the assumption that the prompts are capable of encapsulating the necessary spatio-temporal dynamics. (iii) The final component of the framework involves integrating the text-level embeddings obtained from language model processing with the spatio-temporal attention-transformed time series embeddings derived from time series representation learning. This integration is achieved using the multi-head attention mechanism, which allows for the capture and alignment of contextually relevant information from different domains (textual descriptions and time series data). This fusion leads to more comprehensive and accurate forecasting results, as the model can leverage insights from both textual and numerical data sources. Overall, the novel framework effectively combines the strengths of LLMs, dynamic prompting, time-series-based attention mechanisms, and cross-domain embedding integration to address the challenges of MTSF, resulting in a robust and accurate predictive framework. To evaluate the efficacy and justify the inclusion of each learning component in our framework, we conducted an ablation study. This involved disabling specific components to create various ablated versions, which were then tested on multiple spatio-temporal forecasting datasets in the MTSF task. We compared the performance of our original framework, serving as the baseline, against its ablated variants. These variants, with certain components disabled, exhibited a significant decline in performance, emphasizing the crucial role of each disabled component in the framework. The study employs a comprehensive set of metrics (MAE, RMSE, MAPE), providing an exhaustive evaluation of the ablated variants. The ablation study reinforces the notion that every component is indispensable for the framework to achieve its optimal performance in the MTSF task. The ablated variants that exclude the language model processing, dynamic prompting mechanism, intra-series dependencies, inter-series dependencies, and cross-modal alignment method are labeled as proposed framework `w/o LLMs', `w/o DP', `w/o IntraS', `w/o InterS', and `w/o CMA', respectively; w/o stands for ``without". Specifically, in the case of `w/o CMA', we concatenate the cross-domain embeddings and then apply a linear layer to transform them for label prediction. The ablation study results are presented in Table \ref{tab:abstudy}. A key finding is the superior forecasting performance of the \textbf{LLM-TS Net} and its uncertainty-weighted variant \textbf{w/Unc-LLM-TS Net}, compared to the other ablated models across all datasets (PeMSD3, PeMSD4, PeMSD7, PeMSD8, and PeMSD7(M)). Our ablation study emphasized the critical importance of each component in our framework. Notably, removing specific components from these models leads to a noticeable decline in performance, demonstrating the importance of these elements in the models' overall efficacy. The decline in performance is most pronounced in the ablated variant lacking the cross-modal multi-head attention mechanism (CMA), which exhibits the highest error rates, underscoring its critical role in enhancing forecast accuracy. Indeed, the increase in error can be attributed to the substitution of a simplified linear layer. Furthermore, ablated variants without the dynamic prompting mechanism and language processing also exhibit significant performance drops, highlighting the importance of these features in capturing complex patterns and dependencies within and across time series. Moreover, the performance of the ablated variants varies across different datasets, suggesting that the complexity and characteristics of each dataset uniquely influence the effectiveness of each component in the framework. This variation in performance across datasets indicates that while the \textbf{LLM-TS Net} framework is generally effective, the contributions of its components can vary depending on the dataset's nature. In essence, each component serves a distinct purpose, contributing to a comprehensive strategy for the MTSF task. Their collective inclusion in the framework is warranted by the need for in-depth analysis, flexibility, and a thorough understanding to make precise forecasts. The consistent outperformance of the \textbf{LLM-TS Net} across all datasets underscores its effectiveness.

\vspace{-3mm}
\begin{table*}[ht!]
\centering
\setlength{\tabcolsep}{0.2em} 
\renewcommand\arraystretch{1.14} 
 \resizebox{1.0\textwidth}{!}{
\begin{tabular}{c|ccc|ccc|ccc|ccc|ccc}
\hline
\multirow{2}{*}{\textbf{Methods}} & \multicolumn{3}{c|}{\textbf{PeMSD3}} & \multicolumn{3}{c|}{\textbf{PeMSD4}} & \multicolumn{3}{c|}{\textbf{PeMSD7}} & \multicolumn{3}{c|}{\textbf{PeMSD8}} & \multicolumn{3}{c}{\textbf{PeMSD7(M)}} \\ \cline{2-16} 
 & \multicolumn{1}{l}{\textbf{MAE}} & \multicolumn{1}{l}{\textbf{RMSE}} & \multicolumn{1}{l|}{\textbf{MAPE}} & \multicolumn{1}{l}{\textbf{MAE}} & \multicolumn{1}{l}{\textbf{RMSE}} & \multicolumn{1}{l|}{\textbf{MAPE}} & \multicolumn{1}{l}{\textbf{MAE}} & \multicolumn{1}{l}{\textbf{RMSE}} & \multicolumn{1}{l|}{\textbf{MAPE}} & \multicolumn{1}{l}{\textbf{MAE}} & \multicolumn{1}{l}{\textbf{RMSE}} & \multicolumn{1}{l|}{\textbf{MAPE}} & \textbf{MAE} & \textbf{RMSE} & \textbf{MAPE} \\ \hline
HA & 31.58 & 52.39 & 33.78 & 38.03 & 59.24 & 27.88 & 45.12 & 65.64 & 24.51 & 34.86 & 59.24 & 27.88 & 4.59 & 8.63 & 14.35 \\
ARIMA & 35.41 & 47.59 & 33.78 & 33.73 & 48.80 & 24.18 & 38.17 & 59.27 & 19.46 & 31.09 & 44.32 & 22.73 & 7.27 & 13.20 & 15.38 \\
VAR & 23.65 & 38.26 & 24.51 & 24.54 & 38.61 & 17.24 & 50.22 & 75.63 & 32.22 & 19.19 & 29.81 & 13.10 & 4.25 & 7.61 & 10.28 \\
FC-LSTM & 21.33 & 35.11 & 23.33 & 26.77 & 40.65 & 18.23 & 29.98 & 45.94 & 13.20 & 23.09 & 35.17 & 14.99 & 4.16 & 7.51 & 10.10 \\
TCN & 19.32 & 33.55 & 19.93 & 23.22 & 37.26 & 15.59 & 32.72 & 42.23 & 14.26 & 22.72 & 35.79 & 14.03 & 4.36 & 7.20 & 9.71 \\
TCN(w/o causal) & 18.87 & 32.24 & 18.63 & 22.81 & 36.87 & 14.31 & 30.53 & 41.02 & 13.88 & 21.42 & 34.03 & 13.09 & 4.43 & 7.53 & 9.44 \\
GRU-ED & 19.12 & 32.85 & 19.31 & 23.68 & 39.27 & 16.44 & 27.66 & 43.49 & 12.20 & 22.00 & 36.22 & 13.33 & 4.78 & 9.05 & 12.66 \\
DSANet & 21.29 & 34.55 & 23.21 & 22.79 & 35.77 & 16.03 & 31.36 & 49.11 & 14.43 & 17.14 & 26.96 & 11.32 & 3.52 & 6.98 & 8.78 \\
STGCN & 17.55 & 30.42 & 17.34 & 21.16 & 34.89 & 13.83 & 25.33 & 39.34 & 11.21 & 17.50 & 27.09 & 11.29 & 3.86 & 6.79 & 10.06 \\
DCRNN & 17.99 & 30.31 & 18.34 & 21.22 & 33.44 & 14.17 & 25.22 & 38.61 & 11.82 & 16.82 & 26.36 & 10.92 & 3.83 & 7.18 & 9.81 \\
GraphWaveNet & 19.12 & 32.77 & 18.89 & 24.89 & 39.66 & 17.29 & 26.39 & 41.50 & 11.97 & 18.28 & 30.05 & 12.15 & 3.19 & 6.24 & 8.02 \\
ASTGCN(r) & 17.34 & 29.56 & 17.21 & 22.93 & 35.22 & 16.56 & 24.01 & 37.87 & 10.73 & 18.25 & 28.06 & 11.64 & 3.14 & 6.18 & 8.12 \\
MSTGCN & 19.54 & 31.93 & 23.86 & 23.96 & 37.21 & 14.33 & 29.00 & 43.73 & 14.30 & 19.00 & 29.15 & 12.38 & 3.54 & 6.14 & 9.00 \\
STG2Seq & 19.03 & 29.83 & 21.55 & 25.20 & 38.48 & 18.77 & 32.77 & 47.16 & 20.16 & 20.17 & 30.71 & 17.32 & 3.48 & 6.51 & 8.95 \\
LSGCN & 17.94 & 29.85 & 16.98 & 21.53 & 33.86 & 13.18 & 27.31 & 41.46 & 11.98 & 17.73 & 26.76 & 11.20 & 3.05 & 5.98 & 7.62 \\
STSGCN & 17.48 & 29.21 & 16.78 & 21.19 & 33.65 & 13.90 & 24.26 & 39.03 & 10.21 & 17.13 & 26.80 & 10.96 & 3.01 & 5.93 & 7.55 \\
AGCRN & 15.98 & 28.25 & 15.23 & 19.83 & 32.26 & 12.97 & 22.37 & 36.55 & 9.12 & 15.95 & 25.22 & 10.09 & 2.79 & 5.54 & 7.02 \\
STFGNN & 16.77 & 28.34 & 16.30 & 20.48 & 32.51 & 16.77 & 23.46 & 36.60 & 9.21 & 16.94 & 26.25 & 10.60 & 2.90 & 5.79 & 7.23 \\
STGODE & 16.50 & 27.84 & 16.69 & 20.84 & 32.82 & 13.77 & 22.59 & 37.54 & 10.14 & 16.81 & 25.97 & 10.62 & 2.97 & 5.66 & 7.36 \\
Z-GCNETs & 16.64 & 28.15 & 16.39 & 19.50 & 31.61 & 12.78 & 21.77 & 35.17 & 9.25 & 15.76 & 25.11 & 10.01 & 2.75 & 5.62 & 6.89 \\
STG-NCDE & 15.57 & 27.09 & 15.06 & 19.21 & 31.09 & 12.76 & 20.53 & 33.84 & 8.80 & 15.45 & 24.81 & 9.92 & 2.68 & 5.39 & 6.76 \\ \hline
\textbf{LLM-TS Net} & \textbf{11.98} & \textbf{17.8} & \textbf{9.44} & \textbf{15.78} & \textbf{22.7} & \textbf{8.87} & \textbf{17.78} & \textbf{27.07} & \textbf{7.67} & \textbf{12.79} & \textbf{18.87} & \textbf{6.79} & \textbf{2.16} & \textbf{4.27} & \textbf{5.22} \\
\textbf{W/Unc-LLM-TS Net} & \textbf{11.89} & \textbf{17.58} & \textbf{9.62} & \textbf{15.99} & \textbf{24.04} & \textbf{8.73} & \textbf{17.91} & \textbf{28.35} & \textbf{8.93} & \textbf{13.22} & \textbf{20.46} & \textbf{7.09} & \textbf{2.21} & \textbf{4.33} & \textbf{5.5} \\ \hline
\end{tabular}
}
\vspace{-2mm}
\caption{The table shows forecast error estimations based on model predictions for a 12-step-ahead forecast horizon on benchmark datasets(PeMSD3, PeMSD4, PeMSD7, PeMSD8, PeMSD7(M)). Lower the error better the model performance.}
\label{tab:results1}
\vspace{-2mm}
\end{table*}

\begin{table*}[ht!]
\centering
\setlength{\tabcolsep}{0.2em} 
\renewcommand\arraystretch{1.14} 
 \resizebox{1.0\textwidth}{!}{
\begin{tabular}{c|ccc|ccc|ccc|ccc|ccc}
\hline
\multirow{2}{*}{\textbf{Performance Increase (\%)}} & \multicolumn{3}{c|}{\textbf{PeMSD3}} & \multicolumn{3}{c|}{\textbf{PeMSD4}} & \multicolumn{3}{c|}{\textbf{PeMSD7}} & \multicolumn{3}{c|}{\textbf{PeMSD8}} & \multicolumn{3}{c}{\textbf{PeMSD7(M)}} \\ \cline{2-16} 
 & \multicolumn{1}{l}{\textbf{MAE}} & \multicolumn{1}{l}{\textbf{RMSE}} & \multicolumn{1}{l|}{\textbf{MAPE}} & \multicolumn{1}{l}{\textbf{MAE}} & \multicolumn{1}{l}{\textbf{RMSE}} & \multicolumn{1}{l|}{\textbf{MAPE}} & \multicolumn{1}{l}{\textbf{MAE}} & \multicolumn{1}{l}{\textbf{RMSE}} & \multicolumn{1}{l|}{\textbf{MAPE}} & \multicolumn{1}{l}{\textbf{MAE}} & \multicolumn{1}{l}{\textbf{RMSE}} & \multicolumn{1}{l|}{\textbf{MAPE}} & \textbf{MAE} & \textbf{RMSE} & \textbf{MAPE} \\ \hline
\textbf{LLM-TS Net} & 23.06 & 34.29 & 37.32 & 17.86 & 26.99 & 30.49 & 13.40 & 20.01 & 12.84 & 17.22 & 23.94 & 31.55 & 19.40 & 20.78 & 22.78 \\
\textbf{W/Unc-LLM-TS Net} & 23.64 & 35.11 & 36.12 & 16.76 & 22.68 & 31.58 & 12.76 & 16.22 &  1.48$\downarrow$ & 14.43 & 17.53 & 28.53 & 17.54 & 19.67 & 18.64 \\
\hline
\end{tabular}
}
\vspace{-2mm}
\caption{The table shows the percentage increase in performance of \textbf{LLM-TS Net} and \textbf{W/Unc-LLM-TS Net} across various metrics on different PeMS datasets (PeMSD3, PeMSD4, PeMSD7, PeMSD8, PeMSD7(M)). The performance increase is calculated relative to the next-best baseline models. }
\label{tab:results2}
\vspace{-2mm}
\end{table*}

\vspace{-4mm}
\begin{table*}[ht!]
\centering
\setlength{\tabcolsep}{0.45em} 
\renewcommand\arraystretch{1.25} 
\resizebox{1.05\textwidth}{!}{
\begin{tabular}{c|c|c|c|c|c|c|c|c|c|c}
\hline
\textbf{Datasets} & \textbf{Increase (\%)} & \textbf{RMSE@3} & \textbf{MAE@3} & \textbf{MAPE@3} & \textbf{RMSE@6} & \textbf{MAE@6} & \textbf{MAPE@6} & \textbf{RMSE@12} & \textbf{MAE@12} & \textbf{MAPE@12} \\
\hline
\multirow{2}{*}{\textbf{METR-LA}} & \textbf{LLM-TS Net} & $\textbf{29.74\%}$ & $\textbf{33.59\%}$ & $\textbf{37.86\%}$ & $\textbf{7.44\%}$ & $\textbf{6.35\%}$ & $\textbf{17.71\%}$ & $\textbf{12.10\%}$ & $\textbf{10.47\%}$ & $\textbf{7.81\%}$ \\
                                  & \textbf{W/Unc-LLM-TS Net} & $\textbf{26.95\%}$ & $\textbf{32.44\%}$ & $\textbf{33.63\%}$ & $\textbf{5.29\%}$ & $\textbf{3.34\%}$ & $\textbf{16.46\%}$ & $\textbf{10.71\%}$ & $\textbf{6.40\%}$ & $\textbf{8.63\%}$ \\ \hline
\multirow{2}{*}{\textbf{PEMS-BAY}} & \textbf{LLM-TS Net} & $\textbf{47.58\%}$ & $\textbf{42.97\%}$ & $\textbf{49.25\%}$ & $\textbf{38.29\%}$ & $\textbf{30.82\%}$ & $\textbf{43.38\%}$ & $\textbf{36.57\%}$ & $\textbf{22.58\%}$ & $\textbf{38.67\%}$ \\
                                   & \textbf{W/Unc-LLM-TS Net} & $\textbf{45.35\%}$ & $\textbf{41.41\%}$ & $\textbf{44.74\%}$ & $\textbf{37.47\%}$ & $\textbf{28.93\%}$ & $\textbf{40.85\%}$ & $\textbf{36.11\%}$ & $\textbf{17.20\%}$ & $\textbf{33.64\%}$ \\
\hline
\end{tabular}
}
\vspace{-2mm}
\caption{The table presents the comparative performance improvements of \textbf{LLM-TS Net} and \textbf{W/Unc-LLM-TS Net} in terms of percentage increase across several metrics on the \textbf{METR-LA} and \textbf{PEMS-BAY} datasets. These improvements are measured against the next-best baseline models.}
\label{tab:results3}
\vspace{-1mm}
\end{table*}

\vspace{-1mm}
\begin{table*}[ht!]
\centering
\setlength{\tabcolsep}{0.2em} 
\renewcommand\arraystretch{1.14} 
\resizebox{1.0\textwidth}{!}{
\begin{tabular}{c|ccc|ccc|ccc|ccc|ccc}
\hline
\multirow{2}{*}{\textbf{Model}} & \multicolumn{3}{c|}{\textbf{PeMSD3}} & \multicolumn{3}{c|}{\textbf{PeMSD4}} & \multicolumn{3}{c|}{\textbf{PeMSD7}} & \multicolumn{3}{c|}{\textbf{PeMSD8}} & \multicolumn{3}{c}{\textbf{PeMSD7(M)}} \\ \cline{2-16} 
 & \multicolumn{1}{l}{\textbf{MAE}} & \multicolumn{1}{l}{\textbf{RMSE}} & \multicolumn{1}{l|}{\textbf{MAPE}} & \multicolumn{1}{l}{\textbf{MAE}} & \multicolumn{1}{l}{\textbf{RMSE}} & \multicolumn{1}{l|}{\textbf{MAPE}} & \multicolumn{1}{l}{\textbf{MAE}} & \multicolumn{1}{l}{\textbf{RMSE}} & \multicolumn{1}{l|}{\textbf{MAPE}} & \multicolumn{1}{l}{\textbf{MAE}} & \multicolumn{1}{l}{\textbf{RMSE}} & \multicolumn{1}{l|}{\textbf{MAPE}} & \textbf{MAE} & \textbf{RMSE} & \textbf{MAPE} \\ \hline
\textbf{LLM-TS Net} & \textbf{11.98} & \textbf{17.8} & \textbf{9.44} & \textbf{15.78} & \textbf{22.7} & \textbf{8.87} & \textbf{17.78} & \textbf{27.07} & \textbf{7.67} & \textbf{12.79} & \textbf{18.87} & \textbf{6.79} & \textbf{2.16} & \textbf{4.27} & \textbf{5.22} \\
\textbf{W/Unc-LLM-TS Net} & \textbf{11.89} & \textbf{17.58} & \textbf{9.62} & \textbf{15.99} & \textbf{24.04} & \textbf{8.73} & \textbf{17.91} & \textbf{28.35} & \textbf{8.93} & \textbf{13.22} & \textbf{20.46} & \textbf{7.09} & \textbf{2.21} & \textbf{4.33} & \textbf{5.5} \\ \hline
\textbf{w/o LLMs} & 14.11 & 21.22 & 11.2 & 18.58 & 26.62 & 10.63 & 21.09 & 32.35 & 9.03 & 14.95 & 22.42 & 7.9 & 2.57 & 5.09 & 6.18 \\ \hline
\textbf{w/o DP} & 14.21 & 21.42 & 11.18 & 18.65 & 26.87 & 10.62 & 21.35 & 31.9 & 9.0 & 14.98 & 22.83 & 7.99 & 2.60 & 5.12 & 6.16 \\ \hline
\textbf{w/o IntraS} & 13.67 & 19.73 & 10.79 & 17.5 & 25.69 & 10.09 & 19.81 & 30.27 & 8.63 & 14.45 & 21.56 & 7.66 & 2.45 & 4.71 & 5.84 \\ \hline
\textbf{w/o w/o InterS} & 13.5 & 20.21 & 10.77 & 17.71 & 25.83 & 9.79 & 19.59 & 31.03 & 8.62 & 14.11 & 20.96 & 7.71 & 2.42 & 4.75 & 5.89 \\ \hline
\textbf{w/o CMA} & 15.53 & 22.81 & 12.01 & 19.82 & 28.57 & 11.39 & 22.42 & 35.0 & 9.95 & 16.23 & 23.62 & 8.58 & 2.74 & 5.34 & 6.66 \\ \hline
\end{tabular}
}
\vspace{-2mm}
\caption{The table presents the results of the ablation study on the MTSF task using benchmark datasets. The performance of the ablated variants drops compared to the original framework (\textbf{LLM-TS Net}, \textbf{W/Unc-LLM-TS Net}).}
\label{tab:abstudy}
\end{table*}

\vspace{5mm}
\subsection{Forecasting Uncertainity}
The \textbf{LLM-TS Net} framework employs a supervised learning approach, centralizing its training on minimizing the mean absolute error (MAE). This error metric quantifies the deviation between the model's forecasts, denoted as $\hat{\mathbf{S}}^{t+1}$, and the actual observed data, symbolized as $\mathbf{S}^{t+1}$. The MAE loss function, expressed as $\mathcal{L}_{\text{MAE}}\left(\theta\right)$, is computed using the following formula:

\vspace{-3mm}
\resizebox{0.935\linewidth}{!}{
\begin{minipage}{\linewidth}
\begin{align}
\mathcal{L}_{\text{MAE}}\left(\theta\right) =\frac{1}{\upsilon}\left|\mathbf{S}^{t+1}-\hat{\mathbf{S}}^{t+1}\right| \label{eq:UCE} \nonumber
\end{align} \nonumber
\end{minipage}
}

\vspace{1mm}
Here, $\upsilon$ represents the forecast horizon. The objective in this process is to iteratively refine the model parameters, $\theta$, to attain the lowest MAE loss, thereby significantly enhancing the accuracy of the model’s forecasts. The \textbf{w/Unc-LLM-TS Net}, an extension of the \textbf{LLM-TS Net}, is specifically designed to assess and quantify uncertainties in model predictions. This capability is crucial for augmenting the model's reliability in practical, real-world decision-making scenarios. The \textbf{w/Unc-LLM-TS Net} excels in predicting time-variant uncertainties in point-wise forecasts, extending across multiple future steps. This feature substantially elevates the trustworthiness and dependability of the model's output. The forecasted predictions of the model, denoted as $\hat{\mathbf{S}}^{t+1}$, follow a heteroscedastic Gaussian distribution. The mean of this distribution is given by $\mu_\phi\big(\mathbf{S}^{t}\big)$, and the variance is represented by $\sigma_\phi^2\big(\mathbf{S}^{t}\big)$. Here, $\mathbf{S}^{t}$ signifies the input time series data. The mathematical representation of this relationship is expressed as:

\vspace{-2mm}
\resizebox{0.93\linewidth}{!}{
\begin{minipage}{\linewidth}
\begin{align}
\hat{\mathbf{S}}^{t+1} \hspace{0.5mm} \sim \hspace{1mm}\mathcal{N}\big(\mu_\phi\big(\mathbf{S}^{t}\big), \sigma_\phi^2\big(\mathbf{S}^{t}\big)\big) \nonumber
\end{align}  
\end{minipage} \nonumber
}

\vspace{1mm}
This formulation implies that the predictions are modeled as a normal distribution whose parameters (mean and variance) are not constant but change in response to the input data. The predicted mean and standard deviation for each point in the time series data ($\hat{\mathbf{S}}^{t+1}$), under the heteroscedastic Gaussian distribution, are derived using a specific equation. This equation is as follows:

\vspace{-1mm}
\resizebox{0.945\linewidth}{!}{
\begin{minipage}{\linewidth}
\begin{align}
\mu_\phi\big(\mathbf{S}^{t}\big), \sigma_\phi^2\big(\mathbf{S}^{t}\big) &= f_\theta\big( \text{MHA} (H_{\text{text}}, \mathbf{\tilde{S}}^{t+1})    \big) \nonumber
\end{align} 
\end{minipage}
}

\vspace{2mm}
In this equation, \( f_\theta \) represents a function parameterized by \( \theta \), which processes the output from the cross-modal alignment component, indicated by \( \text{MHA} (H_{\text{text}}, \mathbf{\tilde{S}}^{t+1}) \). Using this input, the network is designed to forecast the mean (represented as \( \mu_\phi\big(\mathbf{S}^{t}\big) \)) and standard deviation (denoted by \( \sigma_\phi^2\big(\mathbf{S}^{t}\big) \)) of a normal distribution. The network's ability to predict these parameters is crucial for generating accurate and reliable forecasts for upcoming time series data points. The maximum likelihood estimate (MLE) for the predicted Gaussian distribution, denoted by $\hat{\mathbf{S}}^{t+1}$, represents the framework's predictions for future estimates. This estimate represents the most likely values for the distribution based on the observed data. The mathematical formulation of this estimate is as follows:

\vspace{-2mm}
\resizebox{0.965\linewidth}{!}{
\begin{minipage}{\linewidth}
\begin{align}
\hat{\mathbf{S}}^{t+1} = \mu_\phi\big(\mathbf{S}^{t}\big) \nonumber
\end{align}  
\end{minipage}
}

In essence, $\mu_\phi\big(\mathbf{S}^{t}\big)$ provides an estimate of the expected value for future predictions ($\hat{\mathbf{S}}^{t+1}$) based on observed data from previous time points, spanning from $t - W$ to $t-1$ (denoted as $\mathbf{S}^{t} = \mathbf{X}^{t-W: t-1}$). Additionally, $\sigma_\phi^2\big(\mathbf{S}^{t}\big)$ quantifies the model's uncertainty in predicting future values for the next $\upsilon$ time steps, starting from the current time point $t$. The Gaussian likelihood of future values, given these parameters ($\mu_\phi\big(\mathbf{S}^{t}\big)$ and $\sigma_\phi^2\big(\mathbf{S}^{t}\big)$), is mathematically expressed as follows:

\vspace{-2mm}
\resizebox{0.895\linewidth}{!}{
\begin{minipage}{\linewidth}
\begin{multline}
\mathcal{N}(\hat{\mathbf{S}}^{t+1};\mu_\phi\big(\mathbf{S}^{t}\big),\sigma_\phi\big(\mathbf{S}^{t}\big))= 
\\ {\frac {1}{\sigma_\phi\big(\mathbf{S}^{t}\big) {\sqrt {2\pi }}}}\hspace{1.5mm} e^{-{\dfrac {1}{2}}\left({\dfrac {\mathbf{S}^{t+1}-\mu_\phi\big(\bar{\mathbf{X}}_{(t - \tau : \hspace{1mm}t-1)}\big) }{\sigma_\phi\big(\bar{\mathbf{X}}_{(t - \tau : \hspace{1mm}t-1)}\big)}}\right)^{2}}  \nonumber
\end{multline}  
\end{minipage} 
}

\vspace{1mm}
The uncertainty modeling framework is designed to minimize the Gaussian negative log-likelihood loss, which is essentially the negative logarithm of the Gaussian probability density function \cite{nix1994estimating}. This optimization focuses on the model's mean and variance estimates. By doing so, it provides a robust basis for comprehensively understanding and quantifying the inherent uncertainty in the predictions made by the model. Applying a logarithmic transformation to both sides of the equation, we obtain the following description:

\vspace{-2mm}
\resizebox{0.91\linewidth}{!}{
\begin{minipage}{\linewidth}
\begin{multline}
\log\ \mathcal{N}(\hat{\mathbf{S}}^{t+1};\mu_\phi\big(\mathbf{S}^{t}\big),\sigma_\phi\big(\mathbf{S}^{t}\big)) 
\\ = \log\left[{\frac {1}{\sigma_\phi\big(\mathbf{S}^{t}\big) {\sqrt {2\pi }}}}\right] + \log \left[e^{-{\dfrac {1}{2}}\left({\dfrac {\mathbf{S}^{t+1}-\mu_\phi\big(\mathbf{S}^{t}\big) }{\sigma_\phi\big(\mathbf{S}^{t}\big)}}\right)^{2}}\right] \\
 = \log\ {\frac {1}{\sigma_\phi\big(\mathbf{S}^{t}\big)}} + \log\ {\frac {1}{{\sqrt {2\pi }}}} -{\frac {1}{2}}\left(\dfrac {\mathbf{S}^{t+1}-\mu_\phi\big(\mathbf{S}^{t}\big) }{\sigma_\phi\big(\mathbf{S}^{t}\big)}\right)^{2} \\ 
 = -\log\ \sigma_\phi\big(\mathbf{S}^{t}\big) + C -{\frac {1}{2}}\left(\dfrac {\mathbf{S}^{t+1}-\mu_\phi\big(\mathbf{S}^{t}\big) }{\sigma_\phi\big(\mathbf{S}^{t}\big)}\right)^{2}  \nonumber
\end{multline}   
\end{minipage} \nonumber
}

\vspace{1mm}
We drop the constant(C) and the Gaussian negative log likelihood loss is described by, 

\vspace{-3mm}
\resizebox{0.90\linewidth}{!}{
\hspace{-8mm}\begin{minipage}{\linewidth}
\begin{multline}
\mathcal{L}_{\text{GaussianNLLLoss}} = 
\\ \sum_{t=1}^{\text{T}_{train}} \left[\frac{\log \sigma_\phi\big(\mathbf{S}^{t}\big)^2}{2}+\frac{\left(\mathbf{S}^{t+1}-\mu_\phi\left(\mathbf{S}^{t}\right)\right)^2}{2 \sigma_\phi\left(\mathbf{S}^{t}\right)^2}\right] \nonumber
\end{multline}   
\end{minipage} \nonumber
}

\vspace{3mm}
where $\text{T}_{\text{train}}$ denotes the time points in the training set. The negative Gaussian log-likelihood is a measure that evaluates how likely the observed data are, given the estimated mean and variance of the Gaussian distribution. A lower value of this measure indicates a more accurate fit of the Gaussian distribution to the observed data, suggesting a more precise representation of the underlying trends. In the \textbf{w/Unc-LLM-TS Net} framework, a Gaussian likelihood function is used to model future values. Within this model, the mean and variance of the Gaussian distribution are computed by the neural network, where the mean represents the predicted future values and the variance reflects the uncertainty associated with these predictions. By minimizing the negative log-likelihood of the Gaussian distribution, the framework effectively identifies the set of model parameters that best capture the data's characteristics and concurrently provide estimates of prediction uncertainty. To put it succinctly, the \textbf{LLM-TS Net} framework minimizes the \textbf{MAE} ( Mean Absolute Error) to identify the optimal set of model parameters for the best fit to the data. In contrast, the \textbf{w/Unc-LLM-TS Net} framework (which is \textbf{LLM-TS Net} with local uncertainty estimation) focuses on minimizing the \textbf{GaussianNLLLoss} (negative log-likelihood of a Gaussian distribution) to quantitatively assess uncertainty.

\vspace{-1mm}
\subsection{BASELINES}
In evaluating the effectiveness of the new neural forecasting models, specifically the \textbf{LLM-TS Net} and \textbf{w/Unc-LLM-TS Net}, for multivariate time series forecasting (MTSF), it is standard practice to compare them against well-known benchmark algorithms. These benchmarks are selected for their prevalent usage in scientific studies and their proven track record on standard benchmark datasets.

\vspace{0mm}
\begin{itemize}
\item HA~\cite{hamilton2020time}, predicts the next value in a time series by calculating the average of observations within a predefined historical window. 
\item ARIMA is a statistical analysis model, designed to handle time series data that is non-stationary. Nonetheless, it has certain limitations. For instance, ARIMA struggles with processing long-term trends and seasonal patterns that evolve over time.
\item VAR(\cite{hamilton2020time}) expands upon the univariate autoregressive (AR) model. It is specifically crafted to understand and analyze the relationships between multiple time series variables for effectively analyzing and predicting the behavior of complex systems.
\item TCN(~\cite{BaiTCN2018}) architecture is designed for multistep-ahead predictions in time series data. This architecture uses a combination of causal convolutions and dilation layers. Causal convolutions enable the TCN to integrate past data effectively, which is crucial for time series analysis. Meanwhile, dilation layers expand the receptive field of the convolutional filters exponentially. This expansion allows the TCN to capture long-range correlations within the MTS data, enhancing its predictive accuracy and effectiveness.
\item FC-LSTM(~\cite{sutskever2014sequence}) employs an encoder-decoder architecture that utilizes LSTM (Long Short-Term Memory) units equipped with peephole connections. This framework has proven to be highly effective in multi-horizon forecasting tasks. It excels in identifying both complex and nonlinear interdependencies among multiple time series variables within MTS data. Notably, it adeptly captures both short-term and long-term relationships among these variables.
\item GRU-ED(~\cite{cho2014grued}) is an encoder-decoder architecture that utilizes GRU (Gated Recurrent Unit) units. This framework is particularly efficient for sequential data processing, especially in multistep-ahead time series prediction. It achieves this by effectively capturing and utilizing information from previous time steps.
\item DSANet(~\cite{Huang2019DSANet}) is designed for forecasting in correlated time series. It employs Convolutional Neural Networks (CNNs) to effectively identify and leverage long-range dependencies within individual time series. Unlike traditional models, DSANet does not depend on recurrent structures to understand temporal relationships in MTS data. Additionally, it incorporates self-attention mechanisms, which are adept at dynamically recognizing inter-dependencies among different time series. This feature is particularly useful for making accurate predictions several steps ahead in MTS data.
\item DCRNN(~\cite{li2018dcrnn_traffic}), is a sophisticated method that combines the concepts of bidirectional random walks on graphs with the integration of graph convolution and recurrent neural networks. This innovative combination is specifically designed to effectively capture both spatial and temporal dependencies in MTS data. A key feature of DCRNN is its encoder-decoder architecture, which is adept at making multistep-ahead forecasts in MTS data. This architecture significantly enhances forecast accuracy compared to traditional forecasting methods. 
\item STGCN(~\cite{bing2018stgcn}) effectively integrates graph convolution with gated temporal convolution. This combination allows the model to efficiently capture the spatial-temporal relationships among multiple time series variables. It is particularly adept at making multi-step-ahead predictions in MTS data.
\item GraphWaveNet(~\cite{wu2019graphwavenet}), innovatively combines a wave-based propagation mechanism with graph representations through dilated causal convolution layers. This design enables the model to learn an adaptive dependency matrix, which is key for capturing spatial-temporal dependencies in time series data. By integrating these elements, GraphWaveNet excels at making multistep-ahead forecasts, effectively understanding the interdependencies among various time series variables. This leads to enhanced forecasting accuracy.
\item ASTGCN(~\cite{guo2019astgcn}), employs an attention-based spatio-temporal graph convolutional network. This design is specifically tailored to discern both inter- and intra-dependencies within time series data, enabling the model to predict future outcomes multiple steps ahead. A key feature of ASTGCN is its ability to efficiently map out spatial and temporal relationships among various time series variables, leveraging attention mechanisms to enhance its predictive accuracy.
\item STG2Seq(~\cite{bai2019STG2Seq}) is designed for multistep-ahead forecasting in MTS data. It employs a dual approach, combining Gated Graph Convolutional Networks (GGCNs) with a sequence-to-sequence (seq2seq) architecture, which is enhanced by attention mechanisms. This structure allows STG2Seq to effectively capture two key aspects: dynamic temporal correlations, which are the changing relationships over time within each time series variable, and cross-channel information, which refers to the correlations among multiple variables. By integrating these elements, the model can comprehensively understand and predict the complex interrelations between multiple time series variables.
\item STSGCN(~\cite{song2020stsgcn})  is designed to make multi-step ahead predictions in MTS data. This is achieved by layering several spatial-temporal graph convolutional layers. These networks are adept at identifying and interpreting both intra- and inter-dependencies within the graph-structured MTS data. By doing so, STSGCN effectively models the intricate relationships that exist among the various time series variables.
\item LSGCN(~\cite{huang2020lsgcn}) enhances multi-step-ahead forecasting in MTS data. It achieves this by incorporating a graph attention mechanism within a spatial gated block. This innovative approach effectively identifies and leverages dynamic dependencies between various time-series variables. It does so through attention mechanisms, which significantly boost the accuracy of its forecasts.
\item AGCRN(~\cite{NEURIPS2020_ce1aad92}) predicts multistep-ahead forecasts in MTS data by employing a method that learns the structure of the graph data adaptively. It is particularly effective in identifying and modeling the complex dependencies and relationships that exist in spatial-temporal data, thanks to its ability to discern both intra- and inter-correlations specific in the time series data. This approach allows for a more nuanced understanding of the interactions between various time series variables, resulting in better forecast accuracy.
\item STFGNN(~\cite{li2021stfgnn}) is designed for multistep-ahead forecasting in MTS data. It enhances prediction accuracy by integrating two distinct components: a temporal graph module and a gated convolution module. These modules function concurrently over various time periods. The temporal graph module focuses on capturing the dynamic relationships across different time steps, while the gated convolution module specializes in understanding spatial relationships. Together, they effectively learn and model the complex interdependencies among multiple time series variables.
\item Z-GCNETs(~\cite{chen2021ZGCNET}) predicts multistep-ahead forecasts in MTS data. This approach uniquely combines a time-aware zigzag topological layer with time-conditioned Graph Convolutional Networks (GCNs). The key innovation here is the ability to uncover hidden spatial-temporal dependencies and acquire significant time-conditioned topological information. This dual focus allows it to effectively represent the intricate interrelations among various time series variables, taking into account their topological characteristics.
\item STGODE(~\cite{fang2021STODE}) is designed for multistep-ahead forecasting. It employs a tensor-based ordinary differential equation (ODE) to effectively grasp the complex interdependencies among multiple time series variables within MTS data. This methodology facilitates the creation of deeper network architectures. By doing so, it comprehensively captures the intricate spatial-temporal dynamics present in the data, leading to a notable enhancement in the accuracy of the forecasts.
\end{itemize}

\vspace{-2mm}
\subsection{EXPERIMENTAL SETUP}
\vspace{0mm}
The traffic-related benchmark datasets were categorized into three distinct, non-overlapping, and exclusive subsets with different ratios: a training set for model learning, a validation set for fine-tuning hyperparameters, and a test set for evaluating the model's performance on new, unseen data. Specifically, the PEMS-BAY and METR-LA datasets were divided using a 70$\%$/10$\%$/20$\%$ split for training, validation, and testing, respectively. In contrast, the remaining datasets followed a 60$\%$/20$\%$/20$\%$ division for these three sets. Prior to training the forecasting models, all time-series variables were standardized to have a mean of zero and a variance of one. The models were evaluated using several accuracy metrics, including Mean Absolute Error (MAE), Root Mean Squared Error (RMSE), and Mean Absolute Percentage Error (MAPE). These metrics were calculated using the original scale of the time-series data, both during model training and evaluation. The hyperparameters of the dynamic prompting mechanism, involve three main parameters: window size (W=12), prompt pool size (M=15), and embedding dimension (d=64). The window Size is crucial for determining the length of the historical data segment in the sliding window technique, impacting its ability to discern spatio-temporal patterns. The prompt pool size affects the diversity of prompts available, influencing the model’s adaptability to various data patterns. Lastly, embedding dimensions determine the complexity and expressiveness of the prompt and time series embeddings. The hyperparameters for Grouped-query Multi-head Attention (GQ-MHA) include critical factors such as the number of groups ($G$=3), which affects the distribution of attention across data segments. The dimensionality of key/value/query projections($d$), which influences the transformation of input data, and the number of attention heads ($H$=4), which dictates the focus on different input aspects. The scaling factor($d_k$=16), typically a fraction of the key/value dimension, stabilizes the attention mechanism. Balancing these parameters is crucial for achieving an optimal trade-off between the model's accuracy in capturing spatio-temporal dynamics and computational efficiency. The  \textbf{LLM-TS Net} framework was trained for 30 epochs with a batch size(b) of 48, an iterative process that refined its parameters by minimizing forecast error. Early stopping based on Validation MAE was implemented to prevent overfitting, ensuring the selection of the best-performing model for effective generalization to unseen data. This strategy enhanced the framework's overall performance. The framework's generalization capabilities were assessed on a test set to evaluate its performance with new data. To improve training efficiency and accelerate model convergence, a learning rate scheduler was implemented. This scheduler dynamically adjusted the learning rate based on the validation set's performance, reducing it by 50$\%$ if there was no improvement for 5 consecutive epochs. This adaptive learning rate strategy optimized the training process and achieved better generalization performance.  Additionally, the Adam optimizer\cite{kingma2014adam} was employed, a widely used optimization algorithm known for its efficiency and robustness in handling large-scale datasets. This optimizer effectively fine-tuned the model's learnable parameters, ensuring smooth and stable convergence. The initial learning rate was set to \num{1e-3}, a carefully chosen value that minimized the MAE loss for the \textbf{LLM-TS Net} model and the negative Gaussian log-likelihood for the \textbf{w/Unc-LLM-TS Net} model. This optimization strategy resulted in model predictions that closely aligned with the actual ground truth. The models were trained on powerful GPUs, including the NVIDIA Tesla V100 to accelerate the training process and facilitate the use of larger models and datasets built upon the PyTorch framework. Multiple independent experimental runs were performed, and the ensemble average was reported to provide a reliable evaluation of the models' performance.

\vspace{-2mm}
\paragraph{Fine-Tuning Llama2-7B model:}
The hyperparameters for parameter-efficient fine-tuning based on the LoRA-AMR technique include: (1) Rank ($r=16$): The rank in the low-rank approximation. It controls the trade-off between model capacity and complexity. A higher rank leads to a more expressive model, but also increases the number of parameters. (2) Alpha (($\alpha$)): A scaling factor in the LoRA technique, typically set to a fraction like  $\frac{1}{r}$. Alpha controls the magnitude of the updates applied to the model parameters. A larger alpha leads to more aggressive updates, which can improve performance but may also lead to instability. (3) LoRA dropout: A dropout rate applied to the low-rank adapters during training. Dropout helps to prevent overfitting and improve generalization. A typical value for LoRA dropout is 0.05. The training configuration comprises a batch size of 16 per GPU for efficient resource utilization, 15 epochs to ensure adequate training and convergence, an initial learning rate of 2e-4 for controlled optimization, a weight decay of 0.001 to prevent overfitting, the AdamW optimizer for adaptive learning rate adjustments, and 4-bit quantization via QLoRA-AMR to facilitate efficient fine-tuning on consumer hardware while maintaining comparable performance.  We employed a supervised fine-tuning approach to train the smaller Llama2-7B model. This approach utilized the LoRA-AMR technique and 4-bit Quantization Bit-Width. The training process involved feeding the model pairs of time series data and their corresponding textual descriptions. The goal was to minimize the cross-entropy loss. We utilized Nvidia V100s (32GB RAM) for fine-tuning the Llama2-7B model using the PyTorch deep-learning library.

\vspace{-1mm}
\subsection{Performance Analysis on Multistep Forecasting at each Horizon}
\begin{figure*}[ht!]
\centering
\hspace*{0cm}\resizebox{0.825\textwidth}{!}{
\subfloat[MAE on PeMSD3]{\includegraphics[width=50mm]{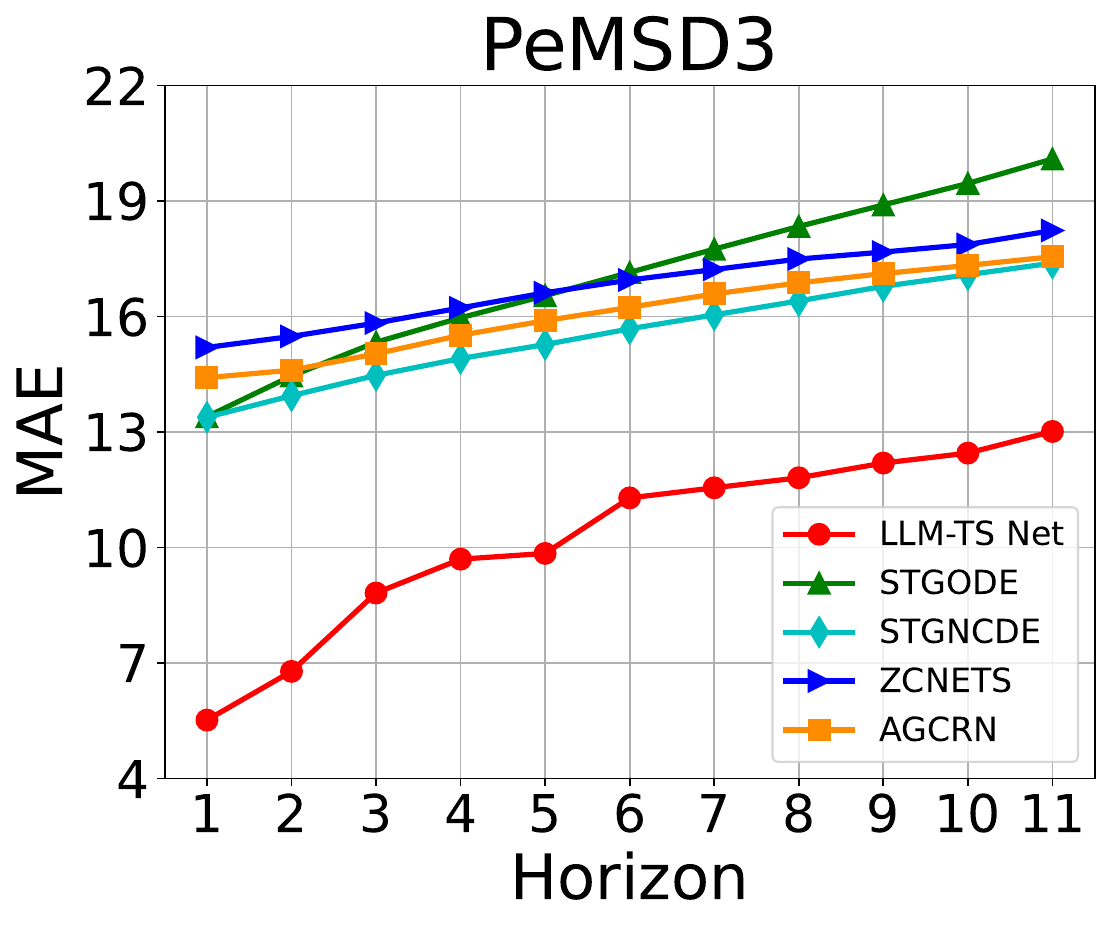}}
\subfloat[MAPE on PeMSD3]{\includegraphics[width=50mm]{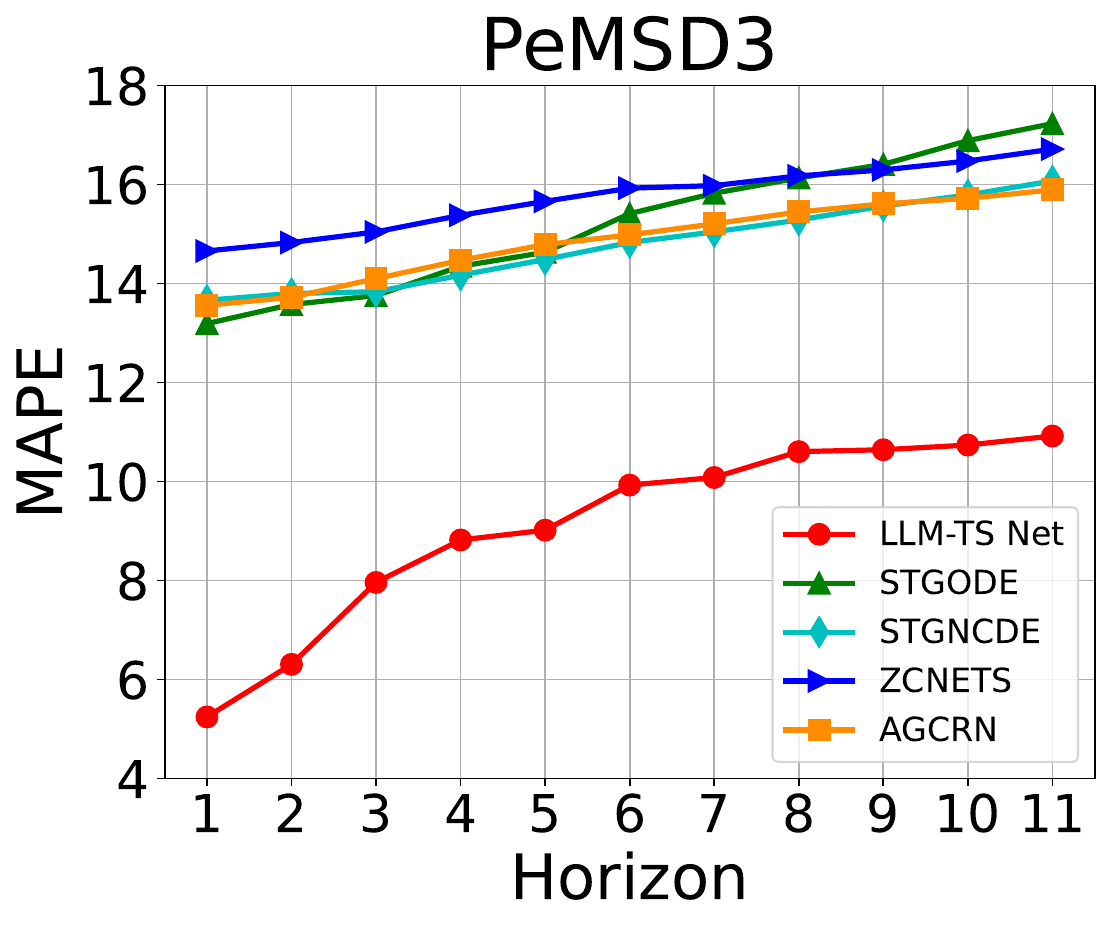}}
\subfloat[RMSE on PeMSD3]{\includegraphics[width=50mm]{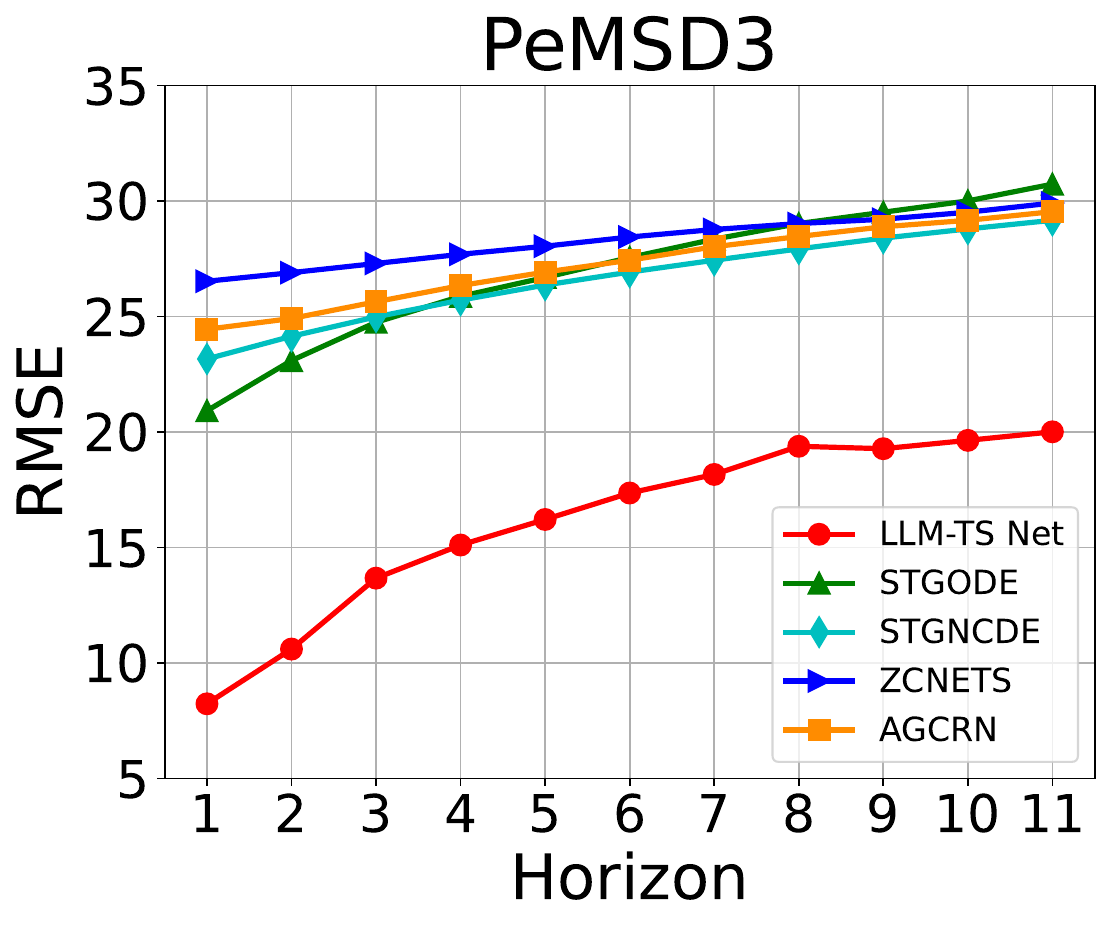}}
}\\[-2ex]
\hspace*{0cm}\resizebox{0.825\textwidth}{!}{
\subfloat[MAE on PeMSD4]{\includegraphics[width=50mm]{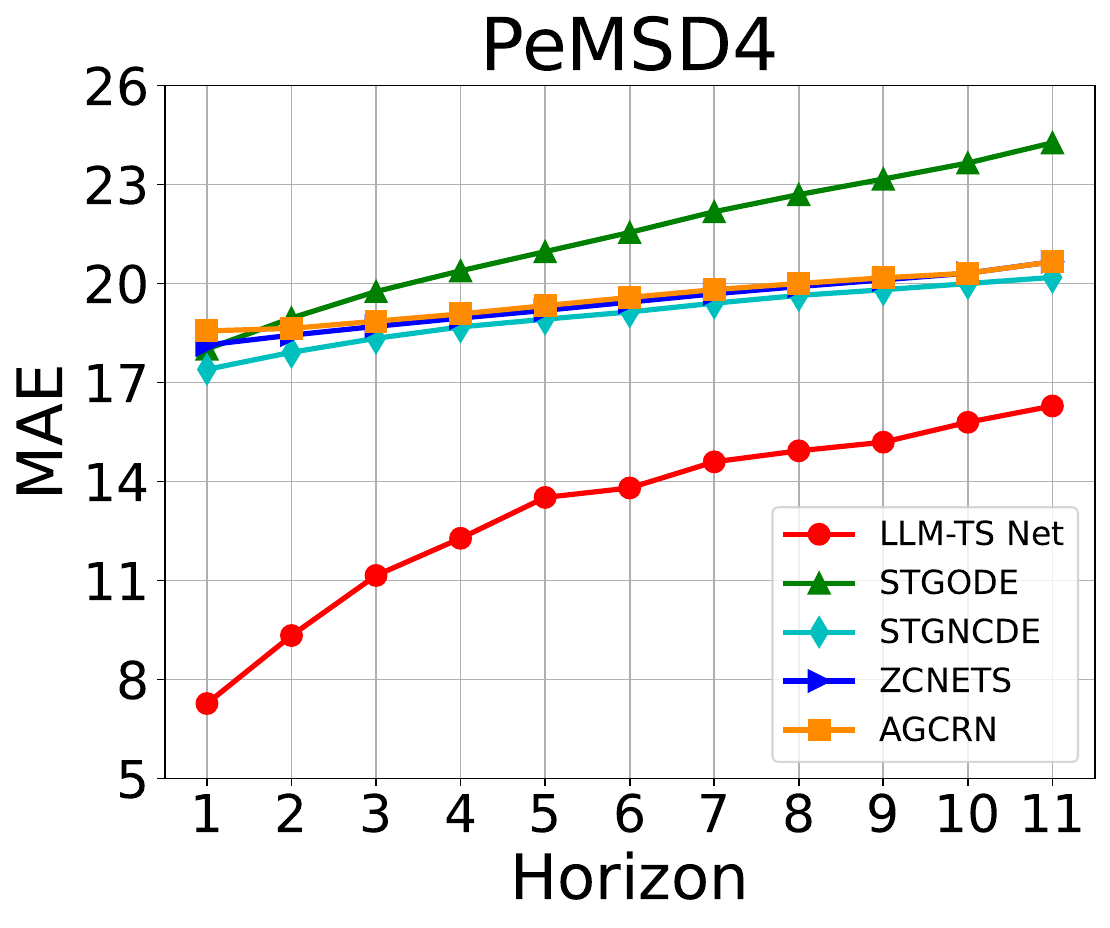}}
\subfloat[MAPE on PeMSD4]{\includegraphics[width=50mm]{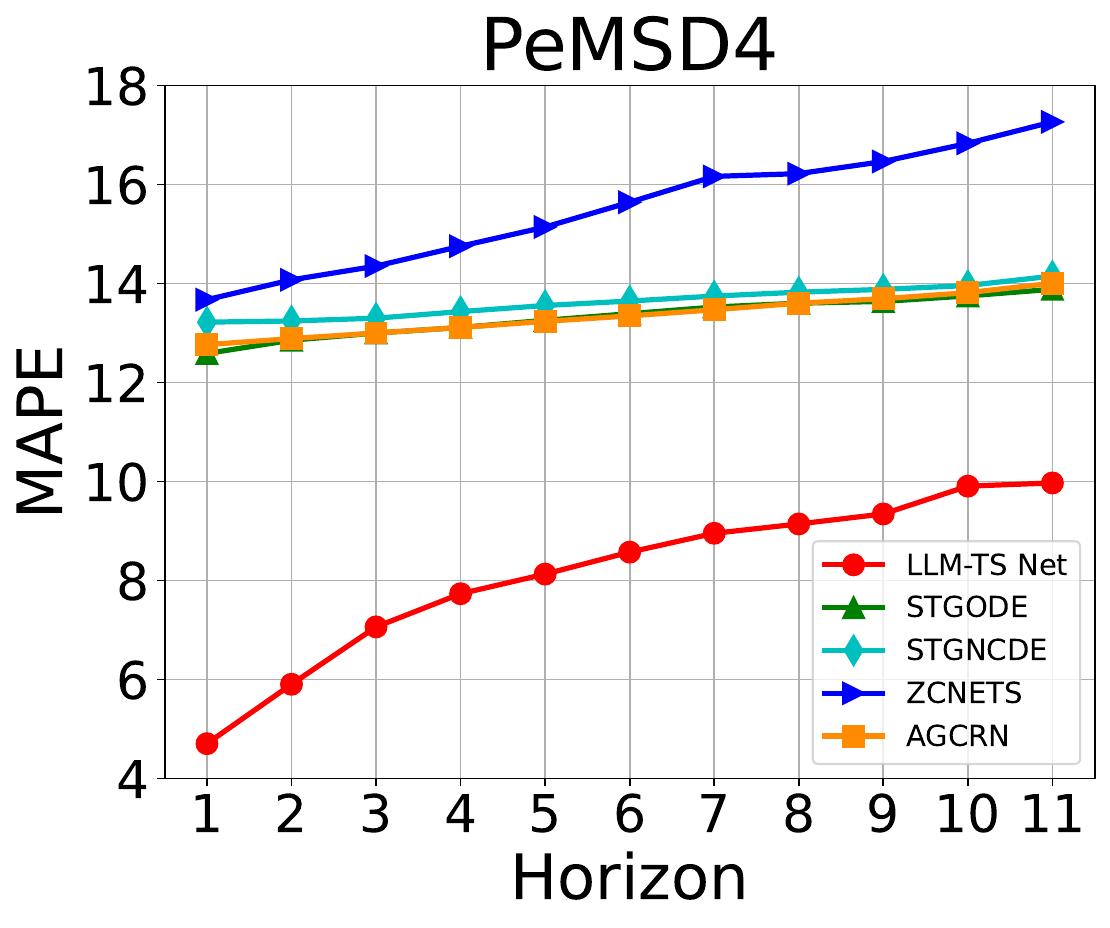}}
\subfloat[RMSE on PeMSD4]{\includegraphics[width=50mm]{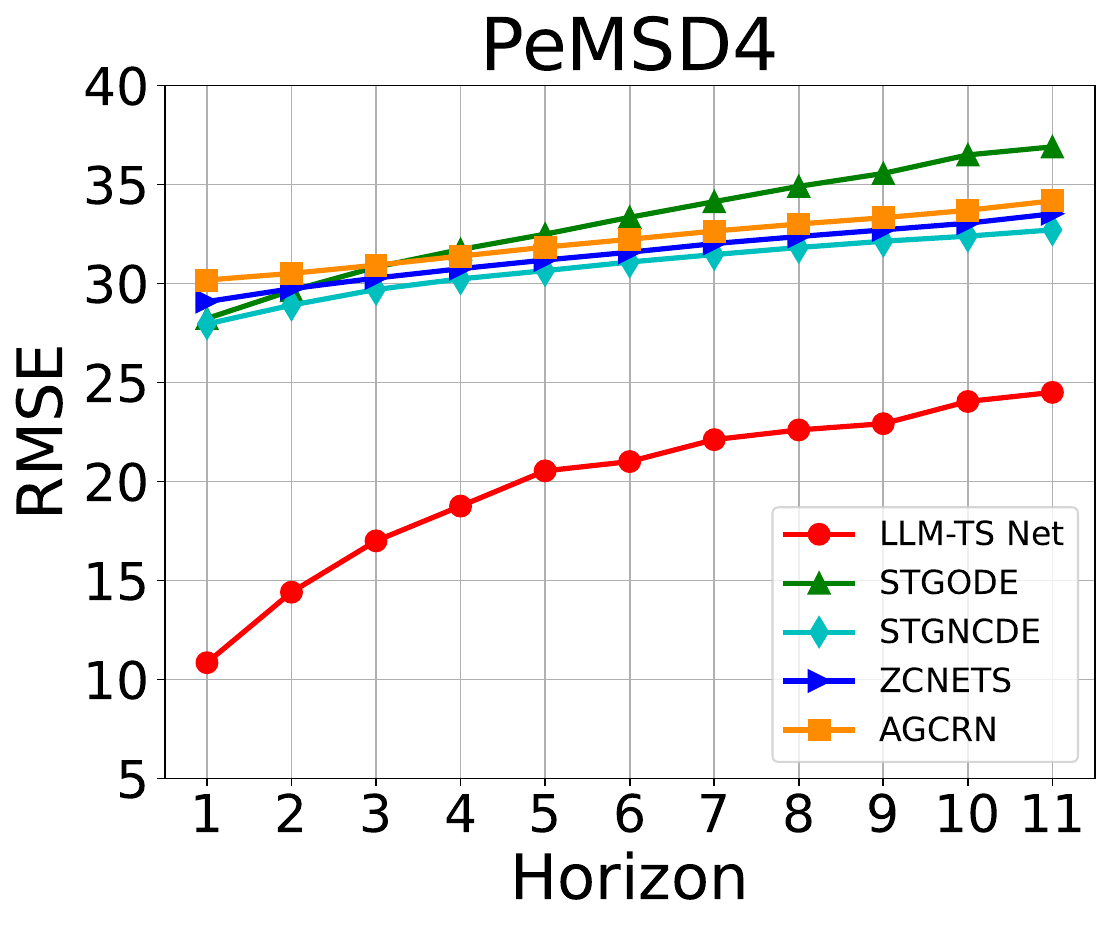}}
}\\[-2ex]
\hspace*{0cm}\resizebox{0.825\textwidth}{!}{
\subfloat[RMSE on PeMSD7]{\includegraphics[width=50mm]{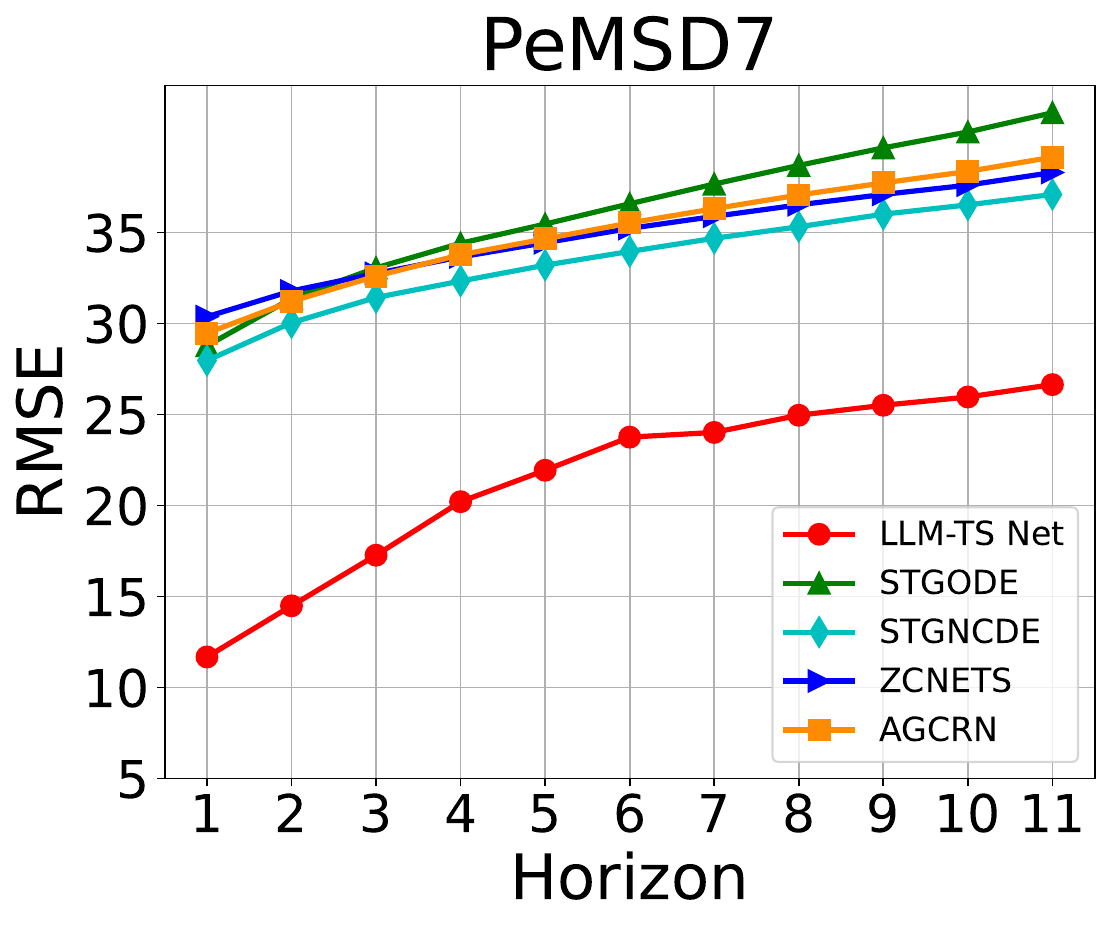}}
\subfloat[MAE on PeMSD7]{\includegraphics[width=50mm]{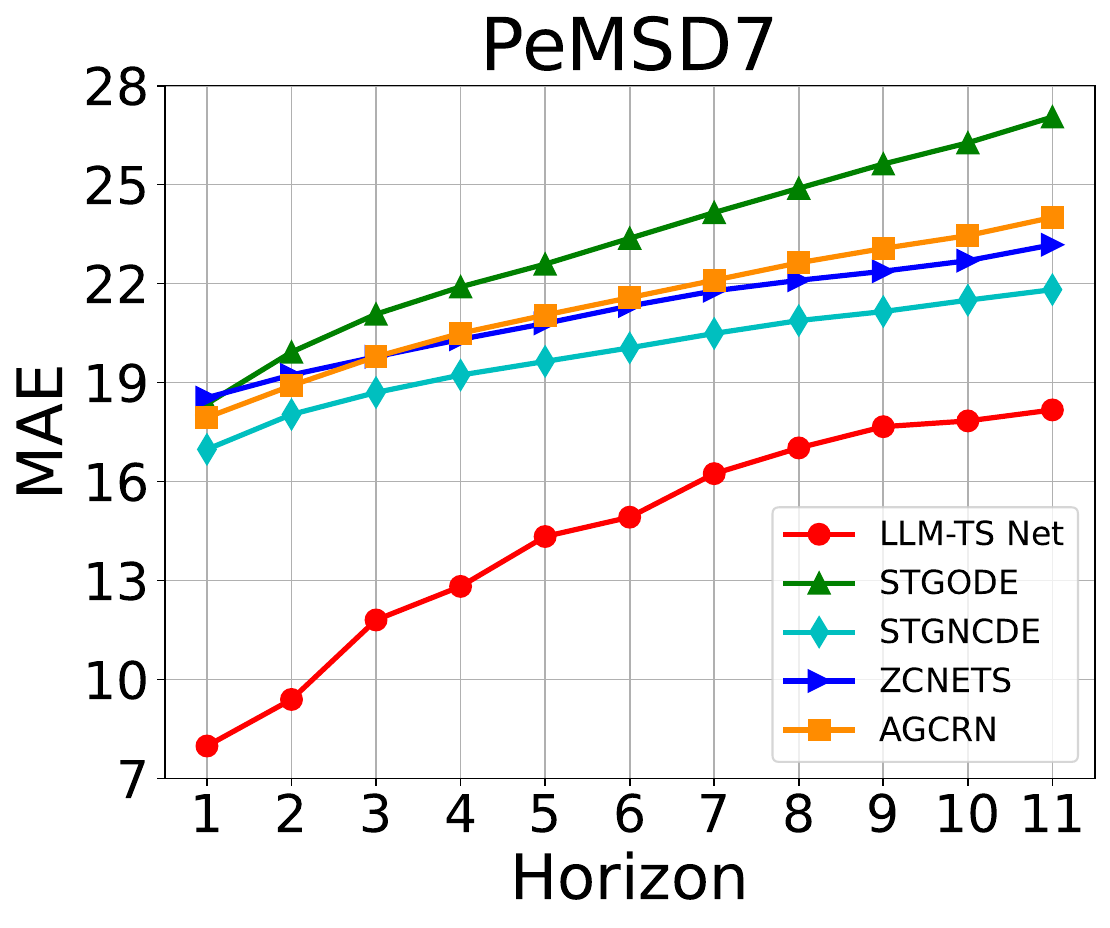}}
\subfloat[MAPE on PeMSD7]{\includegraphics[width=50mm]{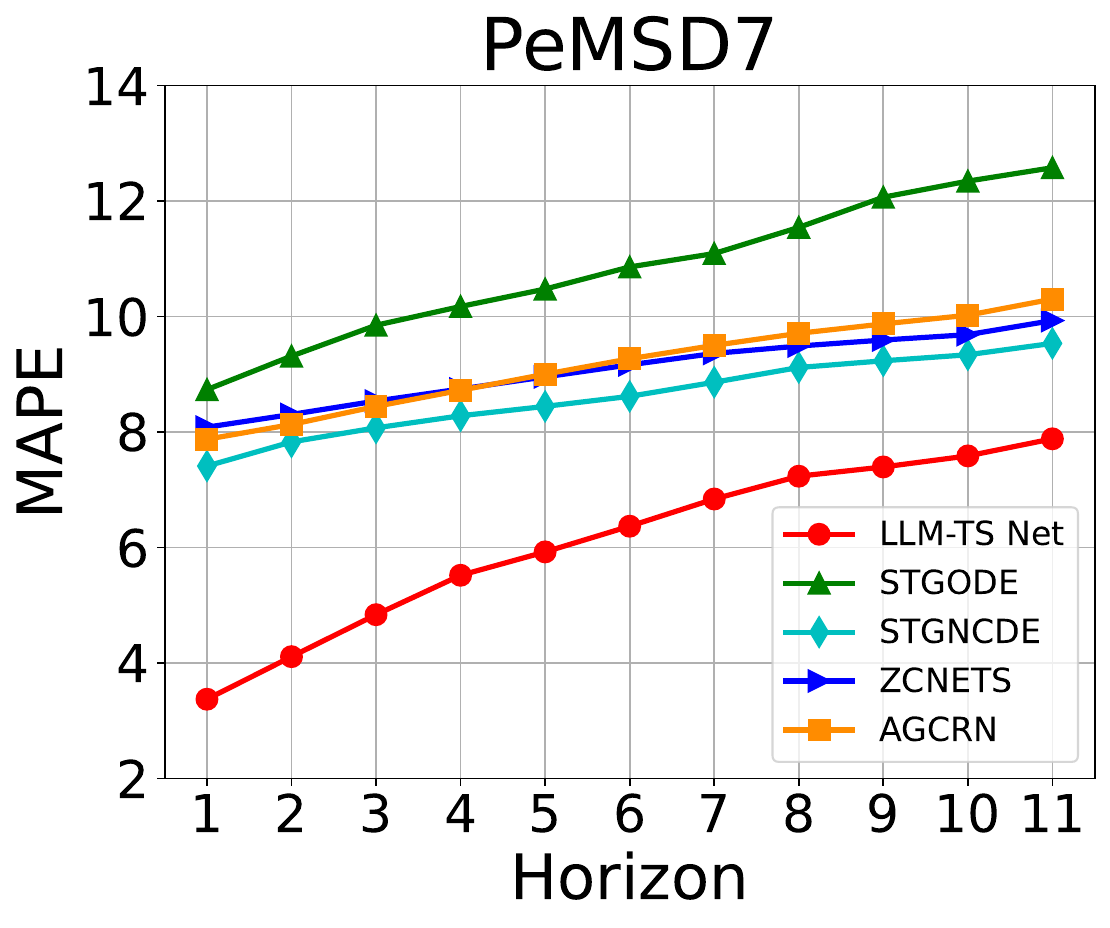}}
}\\[-2ex]
\hspace*{0cm}\resizebox{0.825\textwidth}{!}{
\subfloat[RMSE on PeMSD7(M)]{\includegraphics[width=50mm]{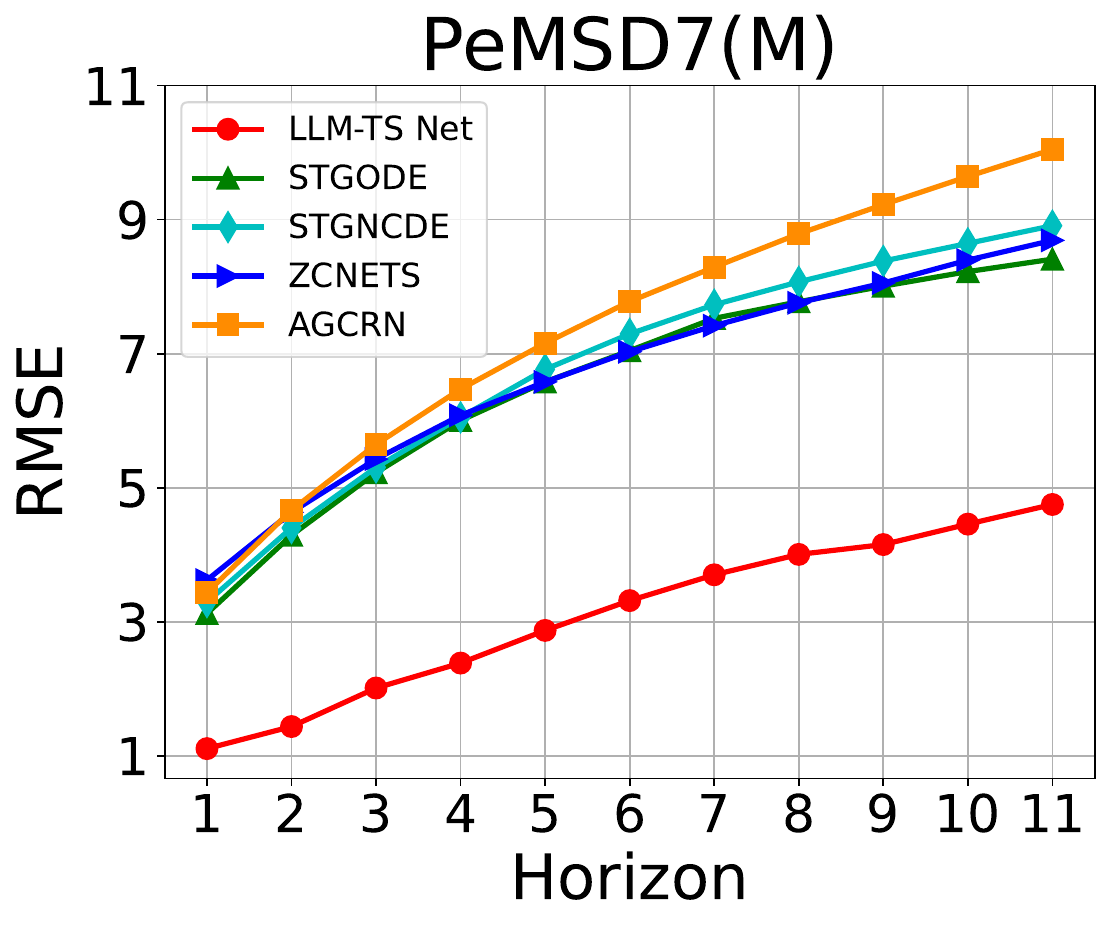}}
\subfloat[MAE on PeMSD7(M)]{\includegraphics[width=50mm]{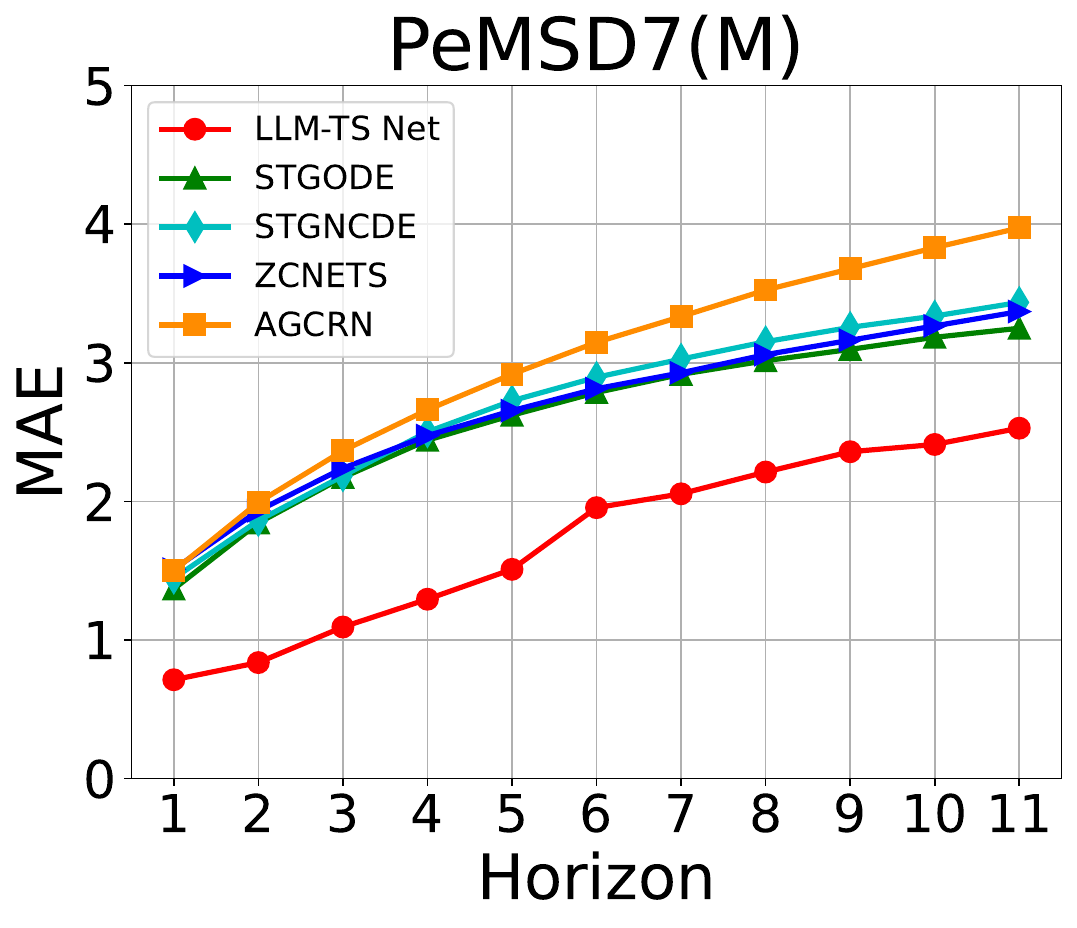}}
\subfloat[MAPE on PeMSD7(M)]{\includegraphics[width=50mm]{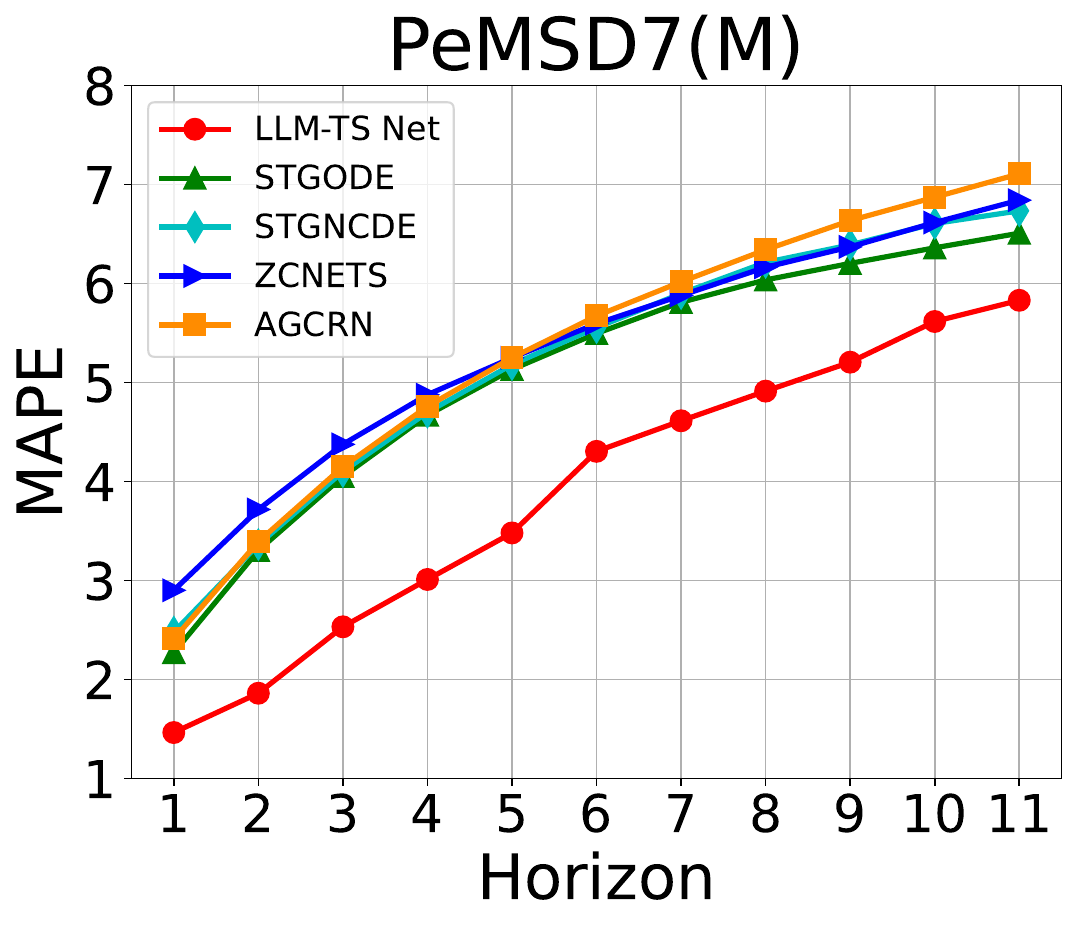}}
}\\[-1ex]
\hspace*{0cm}\resizebox{0.825\textwidth}{!}{
\subfloat[MAE on PeMSD8]{\includegraphics[width=50mm]{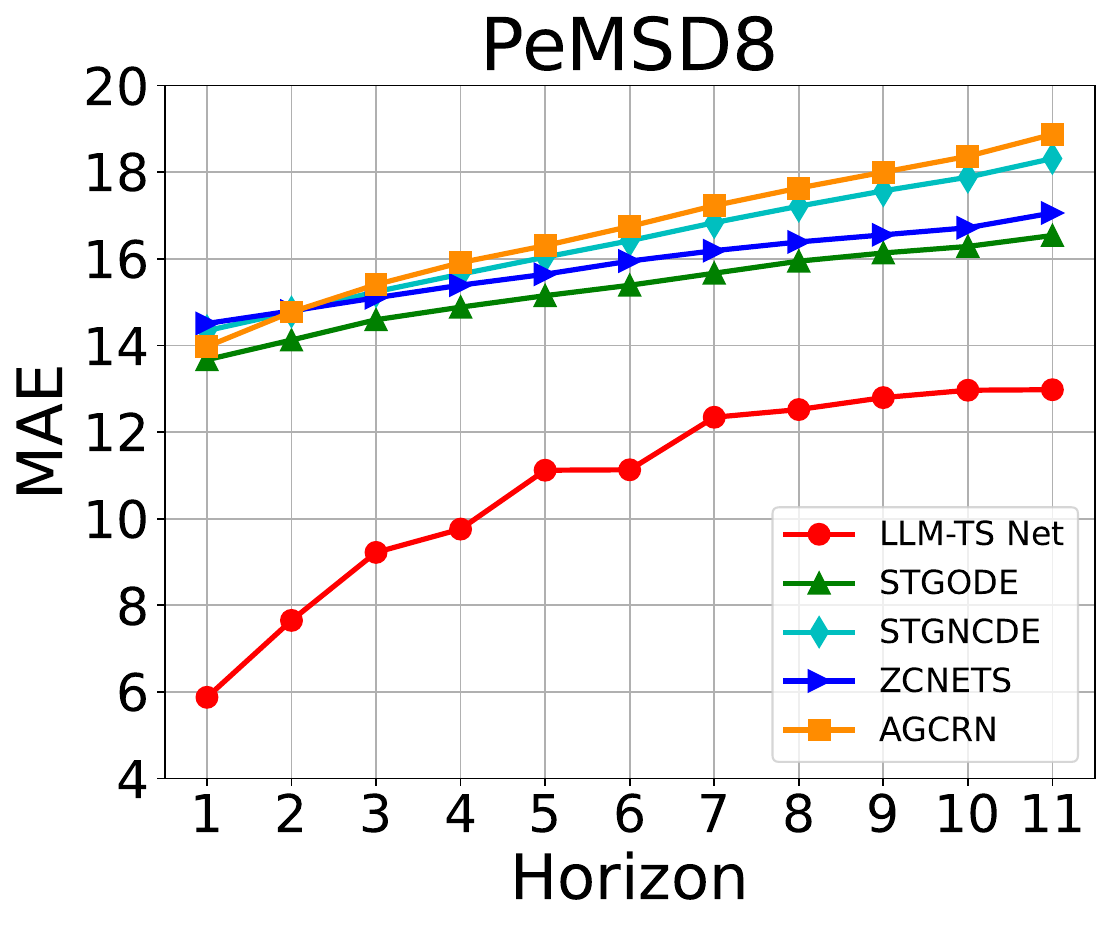}}
\subfloat[MAPE on PeMSD8]{\includegraphics[width=50mm]{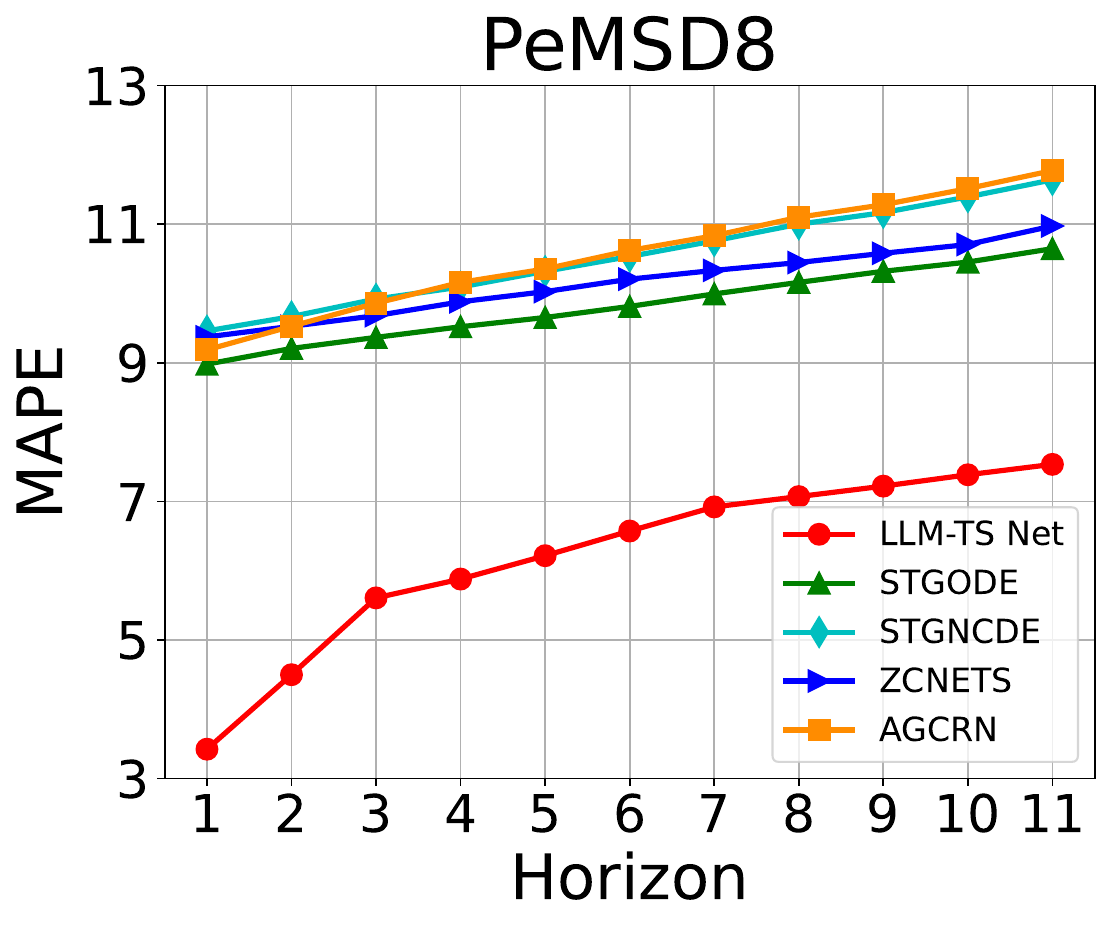}}
\subfloat[RMSE on PeMSD8]{\includegraphics[width=50mm]{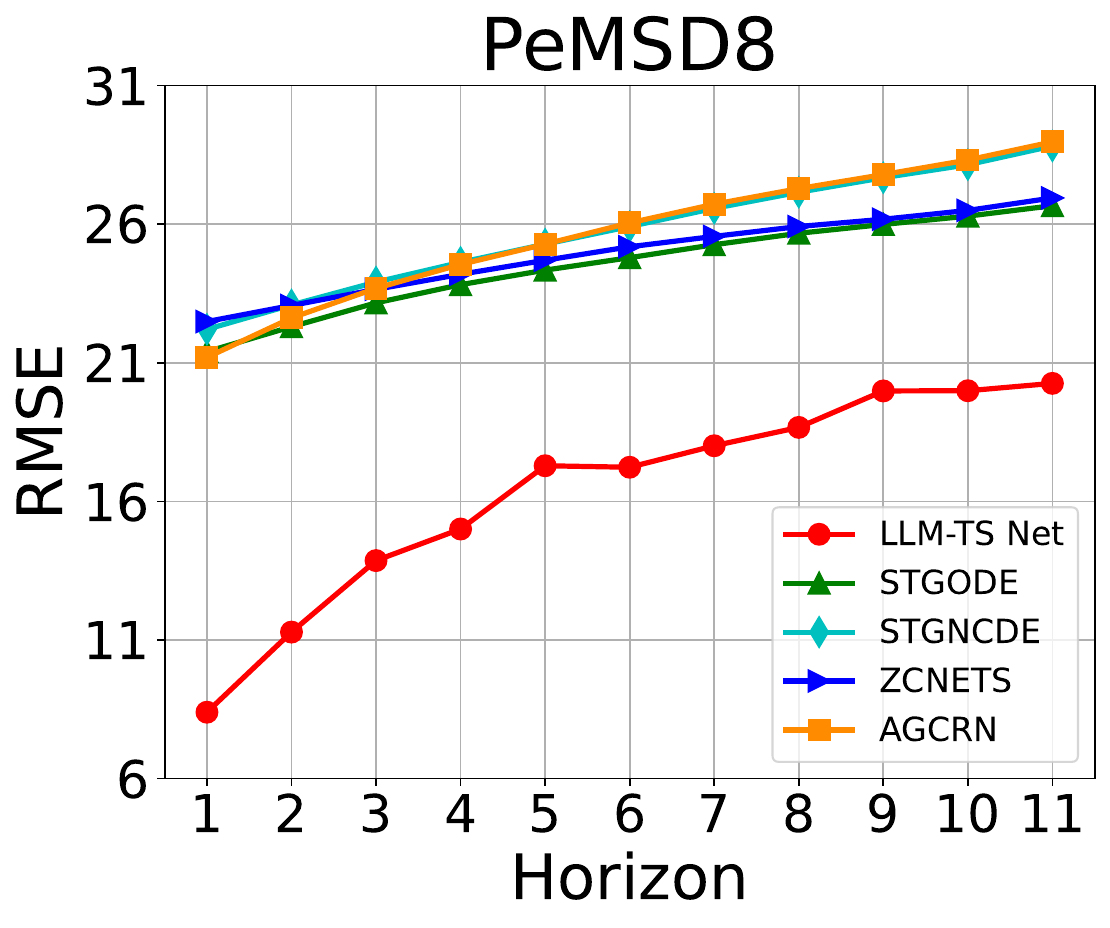}}
}\\[-1.0ex]
\caption{The table shows the pointwise prediction error for multi-horizon forecasting tasks on benchmark datasets.}
\label{fig:ppeh1}
\end{figure*}

The proposed \textbf{LLM-TS Net} neural forecasting framework was evaluated to assess its ability to generate accurate forecasts that extend multiple steps ahead, using a variety of benchmark datasets. Its performance was measured using metrics like Root Mean Square Error (RMSE), Mean Absolute Percentage Error (MAPE), and Mean Absolute Error (MAE), indicating that lower scores denote better performance. Figure \ref{fig:ppeh1} shows the prediction errors for \textbf{LLM-TS Net} multistep-ahead forecasts for various time horizons, compared to other forecasting models, including STGODE, STGNCDE, ZCNETS, and AGCRN, demonstrating their forecasting accuracy for these benchmark datasets. The results demonstrate that \textbf{LLM-TS Net} consistently outperforms the baseline models across all prediction ranges. This finding suggests that \textbf{LLM-TS Net} is highly capable of recognizing and utilizing nonlinear spatio-temporal patterns in MTS data. As the prediction horizon increases, there is a notable rise in forecasting error, as evidenced by the increasing forecasting error values for various models evaluated on the benchmark datasets. This trend highlights the inherent challenge of maintaining accuracy in longer-term predictions.

\vspace{0mm}
\subsection{Irregular Time Series Forecasting}
\vspace{0mm}
Large, complex interconnected sensor networks encompass a diverse range of real-world applications but are plagued by inherent drawbacks of low data quality stemming from inevitable and widespread intermittent sensor failures, faulty sensors, and other factors that arise during the data acquisition process. To assess the effectiveness of the LLM-TS Net framework in dealing with missing data, we simulate data availability and missingness using two different types of missingness patterns \cite{roth2022forecasting, cini2021multivariate} and evaluate the performance of the proposed framework on the missing data for the MTSF task. The simulation techniques mimic missingness patterns that occur in continuous time with asynchronous spatio-temporal patterns commonly encountered in real-world data from large, complex sensor networks. Firstly, the MCAR (Missing Completely At Random) patern simulates sensor failure by randomly dropping observations of each variable within a given historical window, with missing ratios ranging from $10\%$ to $50\%$. In this point-missing pattern, the missingness is unrelated to the observed values or any other variables. This means that the missing data is essentially random and does not provide any information about the underlying process. Secondly, a block-missing pattern refers to a specific arrangement of missing data points in a time series, where consecutive data points are missing for a contiguous period or block. Block-missing patterns are characterized by their length, which represents the number of consecutive missing data points, and their frequency, which indicates how often these patterns occur within the time series. In block-missing patterns with the MCAR (Missing Completely At Random) category, the occurrence of missing blocks is unrelated to the observed values or any other variables. This means that the missingness is essentially random and does not provide any information about the underlying process.
We simulate sensor failure by randomly masking out available data within a given historical window, with missingness ratios ranging from $10\%$ to $50\%$ by selecting appropriate block lengths and frequencies. Our meticulously crafted deep learning-based framework, \textbf{LLM-TS Net}, for time series representation learning incorporates dynamic prompting mechanism, spatial and temporal inference components to effectively process and analyze MTS data, characterized by intricate dependencies and relationships between variables that evolve over time. We delve further into the intricacies of the \textbf{LLM-TS Net} framework, investigating the impact of its individual components on multi-horizon forecasting accuracy in the presence of irregular missing data. To further assess the significance of each component within the \textbf{LLM-TS Net} framework, we conduct a meticulous ablation study on the benchmark model. By selectively removing either the dynamic prompting mechanism, spatial learning component, or temporal learning component, we evaluate the individual contributions of these components to forecasting accuracy under missing data scenarios. These in-depth investigations underscore the robustness and reliability of the \textbf{LLM-TS Net} framework, demonstrating its effectiveness in real-world applications where missing data is a prevalent challenge. To rigorously evaluate the performance of the \textbf{LLM-TS Net} framework, we employ multiple benchmark datasets and meticulously split them into training, validation, and test sets according to a chronological order. For the METR-LA and PEMS-BAY datasets, we adopt a 7:1:2 ratio, while for the remaining datasets, a 6:2:2 ratio is utilized. We then assess the model's effectiveness on simulated data by evaluating its performance against established forecasting error metrics. This comprehensive evaluation process sheds light on the model's ability to handle missingness and irregular intervals in MTS data, providing valuable insights into its resilience across varying levels of missing data. As a benchmark for the MTSF task, we select the \textbf{LLM-TS Net} framework trained on fully observed data (i.e., 0$\%$ missingness), establishing a baseline for comparison. The performance of the proposed models was assessed by measuring forecast errors for a recognized benchmarking task that involves using historical data from 12 time steps to predict estimates 12 time steps into the future. In Tables \ref{tab:pfe1} and \ref{tab:pfe2}, we report the average error in this scenario, which is typically calculated by comparing the predicted values (forecasts) for each of the 12 future time steps with the actual observed values at those time steps. This contrasts with the results presented in Table \ref{tab:results1}, where the model utilizes 12 historical data points to forecast the value at the 12th future time step for computing the forecasting errors. Tables \ref{tab:pfe1} and \ref{tab:pfe2} summarize the irregular-time-series forecasting performance on the benchmark datasets.
While the proposed model exhibits a slight decline in accuracy compared to the benchmark model for lower percentages of missing data, its performance degrades more rapidly as the proportion of incomplete data increases, leading to consistently lower forecast accuracy across all datasets, regardless of the specific pattern of missing data. This resilience to missing data stems from the model's ability to condition pointwise forecasts on the available observations, circumventing the need to rely on imputed values and enabling a more accurate capture of the underlying dependencies and patterns in the MTS data. Additionally, the model effectively captures the nonlinear spatial-temporal dynamic dependencies within the networks of interconnected sensors, generating more reliable out-of-sample forecasts. Furthermore, the results demonstrate that in various missing data scenarios, the ablation model with only the spatial inference component outperforms the one with only the temporal inference component. However, LLM-TS Net, which incorporates both the temporal and spatial learning components, still surpasses all the ablated models. This suggests that the joint optimization of temporal and spatial inference components is crucial for achieving optimal performance in missing data scenarios. The experimental findings indicate that our framework can effectively learn the spatial-temporal dependencies from partial data across various missingness patterns, leading to reduced forecast error.

\newpage
\pagebreak

\begin{table*}[ht!]
\setlength{\tabcolsep}{0.3em} 
\renewcommand\arraystretch{1.35} 
\centering
\resizebox{0.95\textwidth}{!}{
\hspace*{-0.5cm}\begin{tabular}{c|c|ccc|ccc|ccc|ccc}
\hline
\multirow{2}{*}{\textbf{Missing Scheme}} & \multirow{2}{*}{\textbf{Missing Rate}} & \multicolumn{3}{c|}{\textbf{PeMSD3}} & \multicolumn{3}{c|}{\textbf{PeMSD4}} & \multicolumn{3}{c|}{\textbf{PeMSD7}} & \multicolumn{3}{c}{\textbf{METR-LA}} \\ \cline{3-14} 
 &  & \textbf{RMSE} & \textbf{MAE} & \textbf{MAPE} & \textbf{RMSE} & \textbf{MAE} & \textbf{MAPE} & \textbf{RMSE} & \textbf{MAE} & \textbf{MAPE} & \textbf{RMSE} & \textbf{MAE} & \textbf{MAPE} \\ \hline
\textbf{LLM-TS Net} & \textbf{0\%} & 21.03 & 13.64 & 11.30 & 27.04 & 18.02 & 10.31 & 30.05 & 19.52 & 8.08 & 7.53 & 4.56 & 9.47 \\ \hline
\multirow{3}{*}{Point} & 10\% & 21.83 & 13.50 & 11.97 & 26.97 & 18.11 & 12.01 & 31.24 & 21.79 & 8.55 & 7.92 & 4.87 & 8.63 \\
 & 30\% & 21.99 & 13.48 & 12.83 & 31.07 & 20.05 & 11.95 & 30.66 & 22.05 & 9.09 & 8.47 & 5.46 & 9.64 \\
 & 50\% & 22.08 & 15.46 & 12.67 & 32.10 & 20.65 & 13.70 & 32.41 & 21.91 & 10.35 & 8.77 & 5.82 & 9.29 \\ \hline
\multirow{3}{*}{Point (Only Spatial)} & 10\% & 21.41 & 13.30 & 11.69 & 29.91 & 19.23 & 12.06 & 29.71 & 21.26 & 8.66 & 8.49 & 4.66 & 8.56 \\
 & 30\% & 22.30 & 14.57 & 12.61 & 30.78 & 20.58 & 13.26 & 33.26 & 22.57 & 9.94 & 8.51 & 5.26 & 9.43 \\
 & 50\% & 23.22 & 16.13 & 13.94 & 30.82 & 21.97 & 13.45 & 34.64 & 22.47 & 10.02 & 9.03 & 5.74 & 9.61 \\ \hline
\multirow{3}{*}{Point (Only Temporal)} & 10\% & 31.24 & 21.79 & 16.74 & 36.79 & 27.64 & 16.41 & 42.47 & 31.61 & 13.10 & 8.61 & 5.06 & 8.71 \\
 & 30\% & 37.32 & 27.90 & 21.51 & 46.00 & 31.86 & 18.94 & 51.52 & 37.63 & 17.14 & 9.42 & 6.39 & 9.82 \\
 & 50\% & 43.79 & 29.92 & 22.92 & 52.44 & 39.65 & 20.37 & 59.82 & 44.38 & 18.49 & 10.73 & 6.93 & 10.49 \\ \hline
\multirow{3}{*}{Block} & 10\% & 19.87 & 13.70 & 11.82 & 28.27 & 18.39 & 11.52 & 29.84 & 20.65 & 8.39 & 8.40 & 4.52 & 8.86 \\
 & 30\% & 21.14 & 13.96 & 12.65 & 29.59 & 20.34 & 12.43 & 30.85 & 22.33 & 8.78 & 8.73 & 5.05 & 9.57 \\
 & 50\% & 23.01 & 14.72 & 13.88 & 33.18 & 20.88 & 13.81 & 33.04 & 23.21 & 10.21 & 9.47 & 5.56 & 9.39 \\ \hline
\multirow{3}{*}{Block (Only Spatial)} & 10\% & 21.30 & 14.35 & 12.13 & 28.06 & 18.98 & 11.78 & 29.20 & 20.20 & 8.84 & 8.07 & 5.12 & 8.64 \\
 & 30\% & 21.59 & 14.12 & 12.84 & 32.97 & 22.85 & 13.09 & 32.40 & 22.29 & 9.19 & 8.77 & 5.44 & 9.52 \\
 & 50\% & 22.42 & 15.51 & 12.70 & 30.82 & 21.97 & 13.45 & 34.64 & 22.47 & 10.02 & 9.10 & 5.58 & 10.02 \\ \hline
\multirow{3}{*}{Block (Only Temporal)} & 10\% & 30.16 & 21.58 & 17.16 & 38.46 & 27.53 & 15.29 & 44.08 & 32.03 & 13.59 & 8.82 & 5.22 & 9.38 \\
 & 30\% & 38.01 & 26.54 & 20.49 & 44.34 & 34.78 & 20.35 & 52.89 & 37.95 & 16.44 & 9.43 & 5.95 & 9.99 \\
 & 50\% & 47.01 & 34.03 & 24.21 & 50.98 & 40.37 & 22.34 & 62.31 & 46.46 & 20.33 & 11.01 & 6.78 & 10.50 \\ \hline
\end{tabular}
}
\vspace{-2mm}
\caption{Pointwise forecasting error on irregular PeMSD3, PeMSD4, PeMSD7, and METR-LA}
\label{tab:pfe1}
\end{table*}

\vspace{-2mm}
\begin{table*}[ht!]
\setlength{\tabcolsep}{0.45em} 
\renewcommand\arraystretch{1.275} 
\centering
 \resizebox{0.85\textwidth}{!}{
\begin{tabular}{c|c|ccc|ccc|ccc}
\hline
\multirow{2}{*}{\textbf{Missing Scheme}} & \multirow{2}{*}{\textbf{Missing Rate}} & \multicolumn{3}{c|}{\textbf{PeMSD7(M)}} & \multicolumn{3}{c|}{\textbf{PeMSD8}} & \multicolumn{3}{c}{\textbf{PEMS-BAY}} \\ \cline{3-11} 
 &  & \textbf{MAE} & \textbf{RMSE} & \textbf{MAPE} & \textbf{MAE} & \textbf{RMSE} & \textbf{MAPE} & \textbf{MAE} & \textbf{RMSE} & \textbf{MAPE} \\ \hline
\textbf{LLM-TS Net} & \textbf{0\%} & 4.97 & 2.71 & 5.57 & 21.49 & 13.85 & 7.80 & 2.81 & 1.50 & 2.73 \\ \hline
\multirow{3}{*}{Point} & 10\% & 5.16 & 2.97 & 6.85 & 22.61 & 15.91 & 8.67 & 2.82 & 1.60 & 3.20 \\
 & 30\% & 5.60 & 3.63 & 7.63 & 24.67 & 16.96 & 9.48 & 3.04 & 1.77 & 3.45 \\
 & 50\% & 5.93 & 3.69 & 7.90 & 26.96 & 17.50 & 9.90 & 3.16 & 2.01 & 3.44 \\ \hline
\multirow{3}{*}{Point (Spatial Only)} & 10\% & 5.26 & 3.32 & 6.71 & 22.56 & 15.43 & 9.42 & 3.00 & 1.75 & 3.21 \\
 & 30\% & 5.63 & 3.56 & 7.53 & 23.66 & 16.25 & 9.16 & 3.01 & 1.80 & 3.23 \\
 & 50\% & 5.93 & 3.88 & 7.89 & 25.09 & 17.34 & 10.32 & 3.12 & 1.91 & 3.31 \\ \hline
\multirow{3}{*}{Point (Temporal Only)} & 10\% & 5.77 & 3.40 & 7.11 & 32.19 & 22.43 & 13.54 & 3.41 & 1.87 & 3.36 \\
 & 30\% & 6.38 & 3.95 & 8.29 & 38.69 & 28.19 & 16.67 & 3.69 & 2.11 & 3.73 \\
 & 50\% & 6.85 & 4.39 & 9.15 & 44.42 & 32.80 & 18.54 & 3.94 & 2.29 & 3.97 \\ \hline
\multirow{3}{*}{Block} & 10\% & 5.20 & 3.15 & 6.88 & 23.03 & 15.59 & 9.13 & 2.98 & 1.68 & 3.10 \\
 & 30\% & 5.53 & 3.49 & 7.65 & 24.16 & 16.48 & 9.80 & 3.11 & 1.79 & 3.31 \\
 & 50\% & 5.82 & 3.76 & 8.18 & 26.12 & 17.96 & 10.70 & 3.25 & 1.92 & 3.50 \\ \hline
\multirow{3}{*}{Block (Spatial Only)} & 10\% & 5.18 & 3.17 & 6.94 & 23.42 & 15.94 & 9.51 & 2.99 & 1.71 & 3.15 \\
 & 30\% & 5.54 & 3.52 & 7.74 & 24.77 & 17.00 & 10.07 & 3.12 & 1.81 & 3.33 \\
 & 50\% & 5.87 & 3.83 & 8.37 & 26.21 & 18.10 & 10.86 & 3.27 & 1.93 & 3.52 \\ \hline
\multirow{3}{*}{Block (Temporal Only)} & 10\% & 5.77 & 3.39 & 7.04 & 31.95 & 22.18 & 13.32 & 3.40 & 1.86 & 3.34 \\
 & 30\% & 6.36 & 3.95 & 8.21 & 38.57 & 28.11 & 16.63 & 3.69 & 2.10 & 3.72 \\
 & 50\% & 6.90 & 4.42 & 9.08 & 44.82 & 33.10 & 18.50 & 4.00 & 2.35 & 4.02 \\ \hline
\end{tabular}
}\\[-1ex]
\caption{Pointwise forecasting error on irregular PeMSD7(M), PeMSD8 and PEMS-BAY}
\label{tab:pfe2}
\end{table*}

\vspace{-2mm}

\end{document}